\PassOptionsToPackage{HTML,table,dvipsnames}{xcolor}
\PassOptionsToPackage{numbers, compress}{natbib}
\PassOptionsToPackage{section}{placeins}

\documentclass[]{antgroup}
\usepackage{antgroup}

\hypersetup{
  pdftitle={ZoomCap: Evidence-Routed Visual Zooming for Recoverable Visual Information Transfer},
  pdfauthor={Hao Shi, Jiafeng Wang, Yuzhuo Fu, Tao Zhang, Jian Bai, Juelu Zhang, Qing Cui, Chao Huang},
  pdfsubject={Fine-grained image captioning as recoverable visual information transfer},
  pdfkeywords={image captioning, vision-language models, evidence routing, visual zooming, fact-level evaluation}
}

\usepackage{amsmath}
\usepackage{cleveref}
\usepackage{multicol}
\usepackage{amssymb}
\usepackage{array}
\usepackage{tabularx}
\usepackage{bbm}
\usepackage{makecell}
\usepackage{wrapfig}
\usepackage{colortbl}
\usepackage{threeparttable}
\usepackage{pgfplots}
\pgfplotsset{compat=1.18}
\usepackage[normalem]{ulem}
\usepackage{caption}
\useunder{\uline}{\ul}{}

\usepackage{amsmath,amsfonts,bm}

\def\eqref#1{equation~\ref{#1}}

\def\1{\bm{1}}

\DeclareMathAlphabet{\mathsfit}{\encodingdefault}{\sfdefault}{m}{sl}
\SetMathAlphabet{\mathsfit}{bold}{\encodingdefault}{\sfdefault}{bx}{n}

\newcommand{\R}{\mathbb{R}}

\newcommand*\justify{%
  \fontdimen2\font=0.4em
  \fontdimen3\font=0.2em
  \fontdimen4\font=0.1em
  \fontdimen7\font=0.1em
  \hyphenchar\font=`\-
}

\renewcommand{\texttt}[1]{%
  \begingroup
  \ttfamily
  \begingroup\lccode`~=`/\lowercase{\endgroup\def~}{/\discretionary{}{}{}}%
  \begingroup\lccode`~=`[\lowercase{\endgroup\def~}{[\discretionary{}{}{}}%
  \begingroup\lccode`~=`.\lowercase{\endgroup\def~}{.\discretionary{}{}{}}%
  \catcode`/=\active\catcode`[=\active\catcode`.=\active
  \justify\scantokens{#1\noexpand}%
  \endgroup
}

\usepackage{amsmath}
\usepackage{amssymb}
\usepackage{amsfonts}                               
\usepackage{amsthm}
\usepackage[mathcal]{eucal}
\usepackage{mathrsfs}
\usepackage{bm}                                     
\usepackage{blkarray}                               
\usepackage{nicefrac}                               

\usepackage{wrapfig}
\usepackage{graphicx}                               
\usepackage{caption}
\usepackage{cleveref}

\usepackage{tikz}                                          
\usepackage{circuitikz}
\usetikzlibrary{patterns,snakes}
\usetikzlibrary{positioning,calc,fit,decorations.pathmorphing,shapes.geometric, shapes.gates.logic.US, calc}
\usetikzlibrary{arrows,arrows.meta,decorations.markings,shapes,shapes.arrows}
\usetikzlibrary{decorations,decorations.pathreplacing}
\usetikzlibrary{backgrounds}
\usepackage{filecontents}                           
\usepackage{pgfplots}
\usepackage{pgfplotstable}
\usepgfplotslibrary{groupplots}
\usepackage{scalefnt}
\pgfplotsset{compat=newest}

\definecolor{LightGray}{gray}{0.9}
\definecolor{myPink}{HTML}{EE6B98}
\definecolor{myBlue}{HTML}{6BB8FA}
\definecolor{myPurple}{HTML}{9F63F0}
\definecolor{myBlueBase}{HTML}{147BFF}
\definecolor{myGreen}{HTML}{34A853}
\definecolor{myOrange}{HTML}{FF8C00}

\renewcommand\thefootnote{}

\title{ZoomCap: Evidence-Routed Visual Zooming for Recoverable Visual Information Transfer}

\author{\centering
Hao Shi$^{1,2*}$,
Jiafeng Wang$^{1*}$,
Yuzhuo Fu$^1$,
Tao Zhang$^1$,
Jian Bai$^2$,\\
Juelu Zhang$^1$,
Qing Cui$^1$,
Chao Huang$^{1,\dag}$
}

\affiliation{$^1$Ant Group \quad $^2$Zhejiang University}

\begin{document}
\maketitle

\footnotetext{$*$ Equal contribution; $\dag$ Team leader. The controlled ZoomCap-450K corpus is publicly available at \url{https://huggingface.co/datasets/MasterHow/ZoomCap-450K}.}

\renewcommand\thefootnote{\arabic{footnote}}
\setcounter{footnote}{0}


\begin{abstract}

Detailed captions transfer visual information into vision-language model
supervision, but quality is easily conflated with length: long single-pass outputs
still miss small objects, text, and spatial relations or promote uncertain
observations into confident claims. We formulate fine-grained captioning as
\emph{recoverable visual information transfer}, in which a teacher acquires missing
evidence under a finite observation budget and converts it into reusable
supervision. \textbf{ZoomCap} uses a draft caption to route conditional knowledge
retrieval, OCR-sensitive reading, coverage-gap recovery, and adaptive visual
zooming, allocating deeper crops only to the most complex eligible region,
deriving spatial relations from grounded geometry, and fusing non-duplicate
evidence. We also introduce
\textbf{FactCycle}, which scores caption-derived atomic claims on two blind
anchors---source image (truthfulness) and reconstruction (realizability)---separating
fabrication from generator loss claim by claim; human and multi-judge comparisons
complement this fact-level view. Against Qwen3.5-397B-A17B captions, ZoomCap achieves an 81.9\% human tie-adjusted
preference score with a teacher over 10$\times$ smaller, and FactCycle counts 52.4 transferred
claims per image \textit{vs.}\ 31.5 for ScaleCap. The evidence also transfers: in continual pre-training,
ZoomCap-450K improves the 11-benchmark average by 1.31--2.38 points over
the strongest competing corpus and Ling-Tiny-VL-v3 gains 1.4 points on a
production document/OCR monitor suite (1.5 on perception) at 60B tokens; in medical pre-training, 2.64M
ZoomCap-captioned images improve five VQA benchmarks by 3.04 points (\textit{vs.}\ 1.76 for
same-image Gemini-3.1-Pro captions).

\end{abstract}

\newpage

\section{Introduction}

\begin{figure}[t]
\centering
\includegraphics[width=0.9\textwidth]{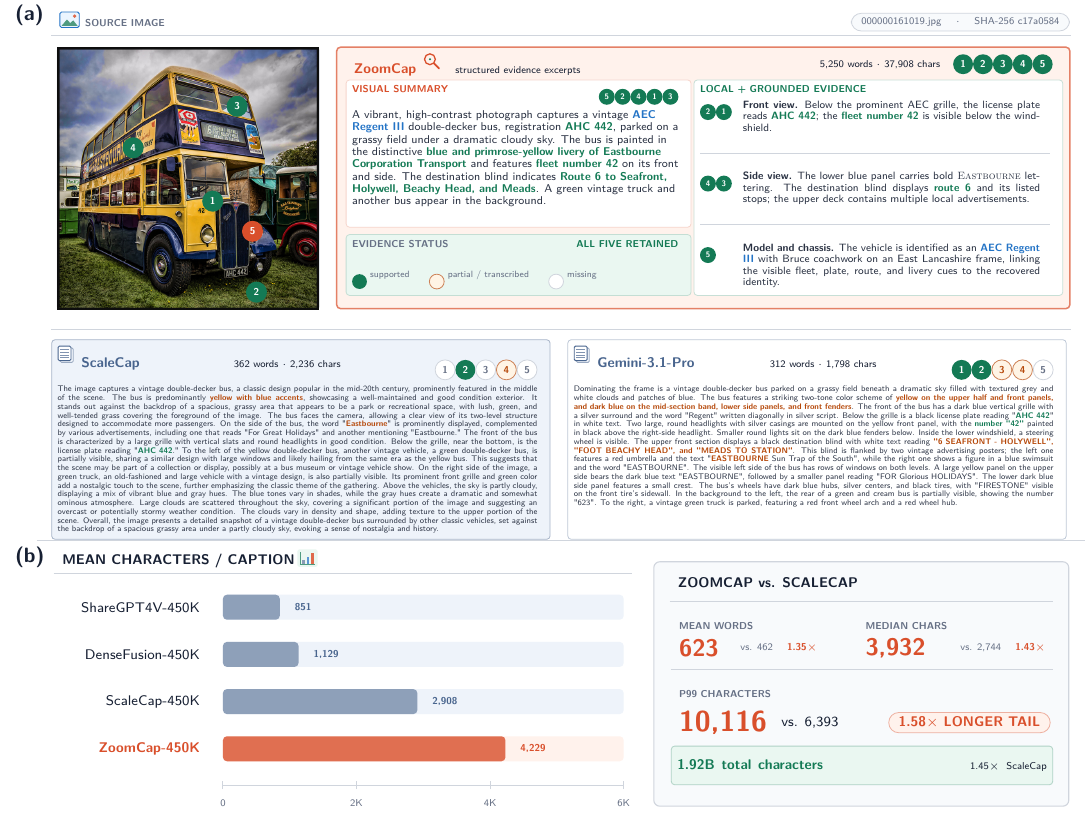}
\caption{\textbf{Evidence preservation comparisons.}
\textbf{(a)} On the same image, five targets remain partial or missing
from ScaleCap~\citep{zhang2025scalecap} and Gemini-3.1-Pro~\citep{google2026gemini31pro} but are retained by ZoomCap. \textbf{(b)} ZoomCap-450K has 1.45$\times$ ScaleCap's total characters
and a 1.58$\times$ larger P99 character length.}
\label{fig:teaser}
\vspace{-8pt}
\end{figure}

In modern VLM training, captions are a textual bottleneck through which visual
information becomes language-model supervision; dense synthetic-caption pipelines
have shown that changing this supervision alone can improve downstream multimodal
models even when architecture and recipe remain fixed~\citep{chen2023sharegpt4v,pi2024imagetextualization,li2024densefusion,awadalla2024blip3kale,zhang2025scalecap,metacaptioner2026}.
The question is therefore whether captions transfer visual evidence that remains
verifiable, recoverable, and useful, not merely whether they are fluent or long.

A single global VLM pass dilutes attention and misses small or peripheral evidence, while fixed high-resolution crops spend the same budget on empty background~\citep{liu2024llavanext,peng2025patchmatters}; retrieval and visual search recover open-world names or overlooked regions~\citep{ramos2023smallcap,li2024evcap,wu2024vstar}, but applying every source to every image is inefficient, and more tokens, tools, or crops need not add trustworthy, geometry-supported information.

This creates a coupled acquisition-and-measurement problem:
generation must decide which evidence source and region justify another
observation and which candidate facts are safe to retain, while evaluation must
quantify what survives once the caption becomes the only interface to the image.
Reference overlap, image--text similarity, and downstream QA on selected
question sets score surface match or backend-specific utility rather than
transferred information~\citep{hessel2021clipscore,dong2024detailcaption,jing2024faithscore,yang2026captionqa};
reconstruction offers a direct test of what survives, yet existing
cycle-consistent metrics collapse it into global, pixel-, semantic-, or
object-level similarity, rewarding layout over individual
content~\citep{ge2024visualfactchecker,cui2025camscore,xu2026itiscore}.
The missing unit is the \emph{traceable visual fact}: which atomic entities,
attributes, relations, counts, small objects, and actions remain realizable
after the caption-to-image round trip---the question FactCycle asks.

We introduce \textbf{ZoomCap}, an agentic teacher for budgeted evidence
acquisition, whose captions retain targets that single-pass engines drop
(Figure~\ref{fig:teaser}):
from a draft caption it conditionally routes world-knowledge search, OCR for
text-structured inputs, and re-observation of draft-omitted regions, expanding
beyond first-level coverage only for the highest-scoring eligible region;
geometry-grounded A-SPOE (attribute-augmented
subject--predicate--object--environment) relations and duplication-avoiding
fusion then merge only
evidence the observations support.

We test this view with FactCycle and downstream transfer. \textbf{FactCycle} extracts
atomic visual claims from the caption text alone and scores the same claim set
with blind judges on two anchors---the source image (truthfulness) and a fixed
text-to-image reconstruction (realizability)---separating what is true from
what is merely renderable, and thus measuring transferred facts rather than
surface similarity. ZoomCap captions contain 67.1 extracted facts per image
(ScaleCap 41.9) and transfer 52.4 of them---$1.7\times$ the strongest
baseline---with higher reconstruction similarity (0.696 vs.\ 0.620), and an
81.9\% human tie-adjusted preference score against single-pass
Qwen3.5-397B-A17B captions:
a teacher over $10\times$ smaller outperforms a one-pass flagship at comparable
per-image cost (Appendix~\ref{sec:efficiency}). Controlled 450K continual pre-training improves the
11-benchmark mean by 1.31--2.38 points across three backbones, and production
adoption (20\% caption-group replacement in a matched 60B-token run; 2.64M
medical images) adds 1.4 document/OCR, 1.5 perception, and 3.04 medical-VQA points (1.28 over
same-image Gemini-3.1-Pro captions).

Our contributions are:
\begin{enumerate}[leftmargin=2em]
    \item We formulate fine-grained captioning as budgeted evidence acquisition and propose ZoomCap, which routes knowledge, OCR, coverage-gap recovery, adaptive zooming, geometry-grounded relations, and draft-anchored gated fusion under a draft caption.
    \item We introduce FactCycle, extending cycle-consistent evaluation from global similarity to traceable atomic claims judged blind on two anchors---source truthfulness and reconstruction realizability---for claim-level caption quality assessment.
    \item We connect direct caption quality to reusable supervision through controlled ZoomCap-450K continual pre-training and production-scale adoption in general and medical domains.
\end{enumerate}


\section{Related Work}
\label{sec:related_work}

\paragraph{Caption supervision and caption-data generation.}
Captions have evolved from a generation target~\citep{xu2015show} into VLM
supervision~\citep{li2020oscar,schuhmann2021laion,li2022blip,li2023blip2,liu2023llava,dai2023instructblip}, and dense synthetic-caption pipelines now use strong MLLMs, vision experts, web knowledge,
or scalable workflows~\citep{chen2023sharegpt4v,pi2024imagetextualization,li2024densefusion,awadalla2024blip3kale,zhang2025scalecap,metacaptioner2026}.
ZoomCap instead routes search, OCR, geometry checks, and deeper crops
conditionally on a grounded draft, spending budget only where the draft remains unexplained---bounded per-image cost with evidence beyond one global pass.

\paragraph{Fine-grained evidence for faithful detailed captioning.}
Region-grounded captioning ties descriptions and relations to local evidence~\citep{johnson2016densecap,krishna2017visualgenome}, while detailed-caption data expose coverage and hallucination risks~\citep{dong2024detailcaption,hua2025finecaption,li2025denseworld}.
High-resolution search, retrieval, and scene graphs recover local details~\citep{liu2024llavanext,peng2025patchmatters,wu2024vstar,ramos2023smallcap,li2024evcap,lu2016visual,xu2017scene,li2024opengraph}.
ZoomCap unifies these channels under one finite observation budget, admitting spatial facts only when geometric margins support them and keeping spatial phrases auditable at full coverage.

\paragraph{Caption evaluation as factual and cycle-consistent measurement.}
Reference-free metrics measure alignment or source-grounded factuality~\citep{hessel2021clipscore,dong2024detailcaption,jing2024faithscore}, and caption-as-visual-encoder protocols test downstream QA~\citep{chen2025goodcaptions,yang2026captionqa}.
Cycle metrics compare reconstructed images globally or by objects~\citep{ge2024visualfactchecker,cui2025camscore,xu2026itiscore}; text-to-image evaluation can decompose prompts into propositions~\citep{hu2023tifa,cho2023dsg}.
FactCycle instead scores caption-derived atomic claims on two blind
anchors---the source image and a fixed reconstruction---separating truthfulness, realizability, and generator loss claim by claim.


\section{ZoomCap Method}\label{sec:pipeline}

\begin{figure}[!hb]
\centering
\includegraphics[width=0.9\textwidth]{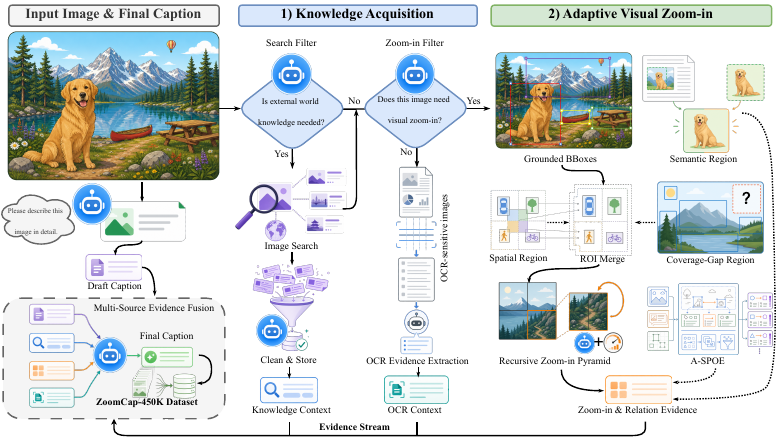}
\vskip -2ex
\caption{\textbf{ZoomCap overview.} From a raw image and a draft caption, ZoomCap routes each sample through agentic evidence acquisition: external-knowledge search, OCR extraction for text-, chart-, and document-like images, and adaptive visual zoom-in for detail-rich scenes. The zoom-in branch combines grounded boxes, semantic regions, caption-coverage gaps, recursive patch observations, and A-SPOE relation facts before multi-source fusion produces the final caption.}
\label{fig:pipeline}
\end{figure}

ZoomCap decouples \emph{what evidence to acquire} from \emph{how to verbalize it} (Figure~\ref{fig:pipeline}): a global draft maps what the image already explains, routes only the necessary evidence streams, and conditions the final fusion.

\subsection{Evidence-Guided Captioning Objective}

Let $I$ be an image and $\mathcal{O}$ its acquired observations: draft $y_0$,
cleaned search context $y_k$, OCR summary $y_{\mathrm{ocr}}$, patch captions
$\{y_p\}_{p\in\mathcal{P}}$, and geometry-derived A-SPOE facts $\mathcal{R}$,
which a draft-anchored gated operator $\mathcal{F}$ fuses into the final
caption $y=\mathcal{F}(y_0,\, y_k,\, y_{\mathrm{ocr}},\, \{y_p\}_{p\in\mathcal{P}},\, \mathcal{R})$.
The teacher should maximize caption utility under a finite observation budget:
\begin{equation}\label{eq:budgeted_captioning}
    y^\star = \arg\max_y U(y; I)
    \quad \textrm{s.t.} \quad
    \sum_{o \in \mathcal{O}} c(o) \le B,
\end{equation}
where $U$ captures coverage, factuality, spatial faithfulness, recoverability, and downstream usefulness. Rather than solving \eqref{eq:budgeted_captioning} exhaustively, a draft-conditioned policy identifies gaps and acquires repair streams; FactCycle (Section~\ref{sec:factcycle_metric}) and the preference/transfer studies instantiate these terms.

\subsection{Draft-Conditioned Evidence Routing}

The first teacher call produces a full-image draft that ZoomCap treats as a scaffold: it supplies the global scene, salient entities, and visible text while exposing what the global view failed to cover, and all routing is conditioned on it.

Three streams repair different gaps. The \emph{knowledge} stream fires when world context unrecoverable from pixels is needed (products, landmarks, artworks, logos, diagrams): image search yields a cleaned context $y_k$ that may name or explain entities but must not override the visual scaffold at fusion. The \emph{OCR} stream serves text-, chart-, document-, and interface-like images, where arbitrary crops destroy reading order, summarizing recognized text into $y_{\mathrm{ocr}}$. The \emph{zoom-in} stream spends additional observations on regions of dense-object, localized-attribute, or spatially structured images that the draft is likely to under-describe.

When zoom-in is selected, ZoomCap extracts localized entities from the draft
and image into $\mathcal{E}=\{(e_i,b_i,a_i)\}_{i=1}^{n}$ (entity name,
normalized box $b_i=[x_1,y_1,x_2,y_2]$, coarse attributes $a_i$); this prior
records what is already grounded, where coverage gaps remain, and which entity
pairs can support geometry-verified relations.

\subsection{Coverage-Aware Visual Zooming}\label{sec:stage3}

\begin{figure}[tb]
\centering
\includegraphics[width=0.9\textwidth]{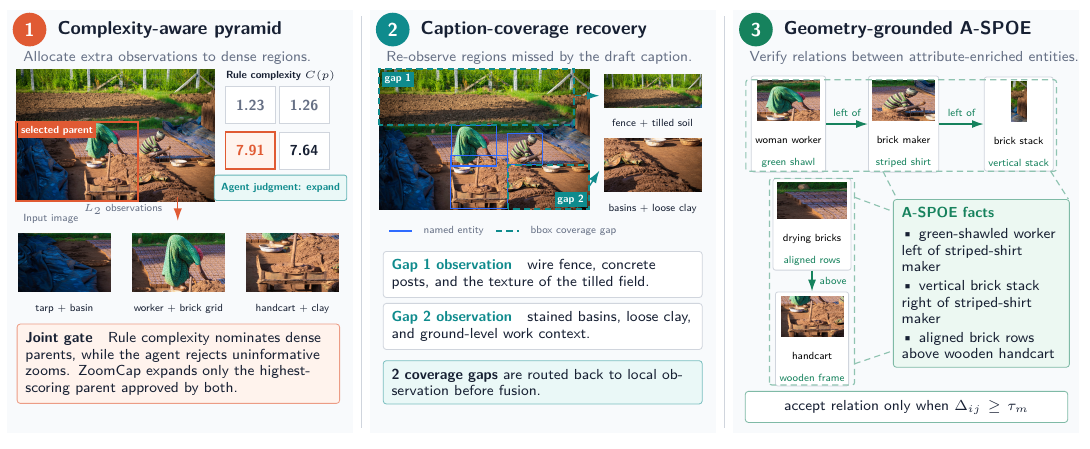}
\vskip -2ex
\caption{\textbf{Adaptive zoom-in mechanisms.} Left: rule complexity nominates dense parents while the agent rejects uninformative zooms; only the highest-scoring parent approved by both expands. Center: caption-coverage gaps derived from draft-grounded entity boxes trigger targeted re-observation (example from PrefBench-456). Right: A-SPOE retains only relations supported by the geometric margin.}
\label{fig:local_mechanisms}
\end{figure}

The zoom-in branch must recover missed evidence without turning every crop into a confident statement: a coverage-aware pyramid uses four quadrants for coarse coverage plus one semantic region when salient content is not quadrant-aligned; for a candidate pool $\mathcal{Q}$ of spatial, semantic, and gap regions, crop selection is
\begin{equation}\label{eq:crop_selection}
    \mathcal{P}^{\star}
    = \arg\max_{\mathcal{P} \subseteq \mathcal{Q},\, |\mathcal{P}| \le K}
    \sum_{p \in \mathcal{P}}\bigl(C(p)+\alpha\, U_{\mathrm{cov}}(p)\bigr)
    - \beta \sum_{p \neq p'} \mathrm{IoU}(p,p'),
\end{equation}
spending observations on complex or draft-under-covered regions while suppressing overlap: quadrants expand around their assigned boxes, the semantic crop groups up to three salient entities, regions merge with grounded boxes and gaps by IoU, and near-duplicates are removed (Figure~\ref{fig:local_mechanisms}).

\paragraph{Caption-coverage gap (CCG) recovery.}\label{sec:ccg}
Let $\Omega_{\mathcal{E}}=\bigcup_i b_i$ be the union of grounded boxes already
explained by the draft caption. Gap recovery scores a candidate region $q$ by
its uncovered fraction $U_{\mathrm{cov}}(q)=|q \setminus \Omega_{\mathcal{E}}|/|q|$,
restricted to regions with $|q|\le\eta_{\max}$ and aspect ratio
$\mathrm{ar}(q)\le\eta_{\mathrm{ar}}$, and selects the largest qualifying gap
rectangles (construction rules in Appendix~\ref{sec:impl_details}).
Unlike generic saliency, $U_{\mathrm{cov}}$ is relative to what the draft
already explains, not to pixel-level conspicuity.

\paragraph{Complexity-gated recursive zooming.}\label{sec:patch_pyramid}
Each first-level patch is captioned independently and retained with its source
coordinates. A rule-based complexity score,
\begin{equation}\label{eq:complexity}
    C(p)=w_{\mathrm{ent}}N_{\mathrm{ent}}(p)+w_{\mathrm{small}}N_{\mathrm{small}}(p)
    +w_{\mathrm{clip}}\rho_{\mathrm{ent}}(p)+w_{\mathrm{area}}r_{\mathrm{area}}(p),
\end{equation}
over entity count, small-object count, clipped entity density, and area ratio
under empirically set non-negative weights
(Appendix~\ref{sec:impl_details}), nominates dense candidates for a second
observation, and an agent judgment vetoes the nominations it deems
uninformative; only the highest-scoring parent approved by both expands, and
that parent with its children passes to fusion. This gating cuts total patch
inputs by 20.2\% and second-level inputs by 39.8\%
(Table~\ref{tab:ablation}(b)).

\subsection{Geometry-Grounded A-SPOE Relations}\label{sec:aspoe}

A-SPOE (attribute-augmented subject--predicate--object--environment) relations
take their predicate-level motivation from relation-aware
captioning~\citep{long2025racap} and add two ingredients of our own: entity
attributes carried on every node, and a geometric admission test that grounds
each accepted spatial predicate in localized boxes. Each grounded entity
becomes a node $v_i=(e_i,a_i,b_i)$ in a graph whose edges store accepted spatial
relations and whose context node stores the scene, making spatial phrases
auditable before they enter free-form text. For an entity pair $(i,j)$, let
$g_{ij}^{x}$ and $g_{ij}^{y}$ be the horizontal and vertical center gaps
normalized by the pair's mean box size; a candidate relation is retained only
when its dominant geometric gap clears an explicit margin,
\begin{equation}\label{eq:aspoe_margin}
    r_{ij}^{\star}=\mathrm{dir}(b_i,b_j)\ \text{if}\
    \max(g_{ij}^{x},g_{ij}^{y}) \ge \tau_{g},\
    \tfrac{\max(g_{ij}^{x},g_{ij}^{y})}{\min(g_{ij}^{x},g_{ij}^{y})} \ge \tau_{r},\
    \mathrm{IoU}(b_i,b_j) \le \tau_{i};\quad \varnothing\ \text{otherwise},
\end{equation}
with $\mathrm{dir}(\cdot)$ from the dominant axis (left/right or above/below).
Appendix~\ref{sec:impl_details} instantiates thresholds, ranking,
deduplication, and per-object caps, so spatial phrases stay grounded in box
geometry rather than left to free-form generation.

\subsection{Draft-Anchored Gated Fusion}\label{sec:fusion}

Routing spends the budget; fusion decides which acquired evidence enters the
caption, as statement-level admission control rather than concatenation: the
draft remains the backbone, and a candidate statement $s$ from any stream is
admitted only when
\begin{equation}\label{eq:fusion_gate}
    g(s)=\mathbf{1}\!\left[\mathrm{Novel}(s;\,y_0,y_k,y_{\mathrm{ocr}})\ge\tau_n
    \,\land\, \mathrm{Consistent}(s;\,y_0)\ge\tau_c
    \,\land\, \max_{s'\in\mathcal{S}_{\mathrm{acc}}}\mathrm{Overlap}(s,s')\le\tau_o\right],
\end{equation}
i.e.\ novel relative to everything already verbalized, consistent with the
visual scaffold, and non-redundant with accepted statements. The gates keep
fusion aligned with \eqref{eq:budgeted_captioning}: retained statements must
increase transferred information, so uncertain local observations never become
confident global claims, and admitted statements keep their provenance instead
of dissolving into one opaque rewrite. They are realized training-free at
prompt level (Appendix~\ref{sec:impl_details}).

\subsection{FactCycle: Two-Anchor, Fact-Level Measurement}\label{sec:factcycle_metric}

\begin{figure}[t]
\centering
\includegraphics[width=\textwidth]{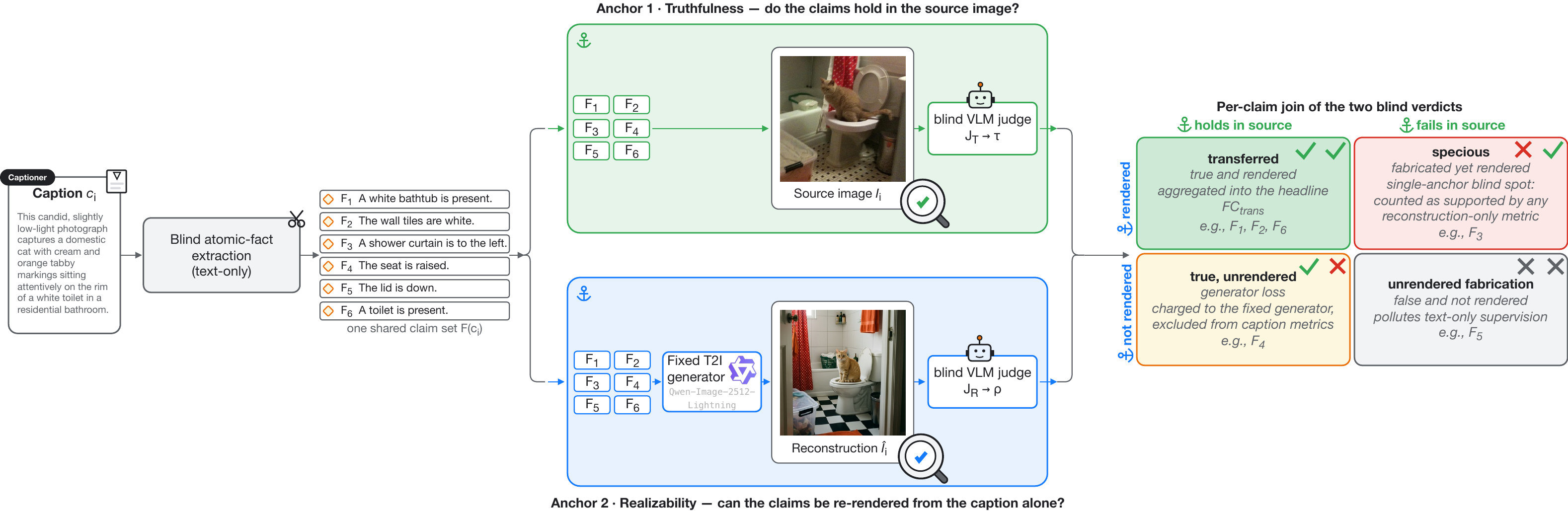}
\caption{\textbf{FactCycle under its two blind anchors.} The same blind-extracted claim set $\mathcal{F}(c_i)$ is scored against the source image (Truthfulness $\tau$) and against the fixed-generator reconstruction (Realizability $\rho$); joining the two verdicts per claim yields four cells, and the headline $\mathrm{FC_{trans}}$ aggregates the transferred cell.}
\label{fig:factcycle_dual_anchor}
\end{figure}

Existing caption-to-image cycle metrics summarize reconstructions through
global features, pixel differences, or detected
objects~\citep{ge2024visualfactchecker,cui2025camscore,xu2026itiscore}, judging
only what the reconstruction shows; FactCycle instead scores the \emph{same}
blind-extracted claims against two blind anchors
(Figure~\ref{fig:factcycle_dual_anchor}): do they hold in the source, and do
they re-render from the caption alone?
A blind text-only extractor turns a caption $c_i$ into atomic visual claims
$\mathcal{F}(c_i)$ (entities, attributes, relations, counts, small objects,
actions, and a residual other bucket); a fixed generator reconstructs $\hat I_i$
from $c_i$ alone.
An independent judge pair---two VLMs, neither of which produces the
reconstructions---applies an identical supported/unsupported/unverifiable rubric
to the anonymized claims on each anchor: $J_T$ sees only $I_i$, $J_R$ only
$\hat I_i$; neither sees the caption or the other anchor's verdicts.
Abbreviating $\mathrm{supported}$ as $\mathrm{sup}$, let $a_f,b_f\in[0,1]$ be
the blind support scores of claim $f$ on the two anchors, i.e.\ the mean over
that anchor's judge pair of $\mathbf{1}[J_T(I_i,f)=\mathrm{sup}]$ and of
$\mathbf{1}[J_R(\hat I_i,f)=\mathrm{sup}]$ (a unanimous pair reduces them to
single-judge indicators), and let
$\mathcal{D}_T(i)$, $\mathcal{D}_R(i)$,
$\mathcal{D}(i)=\mathcal{D}_T(i)\cap\mathcal{D}_R(i)$ be the claims each
anchor, or both, finds decidable on image $i$.
Truthfulness is the source-row marginal,
\begin{equation}\label{eq:anchors}
\tau_i=\frac{1}{|\mathcal{D}_T(i)|}\sum_{f\in\mathcal{D}_T(i)} a_f,
\end{equation}
averaged over images; Realizability $\rho$ is the analogous supported
share over $\mathcal{D}_R(i)$.
Because the anchors share one claim set, the products of $(a_f,b_f)$ with
their complements split every claim of $\mathcal{D}(i)$ into four cells that
sum to one; we report three of them as caption-quality rates: \emph{transferred} (true and rendered), \emph{specious} (fabricated yet
rendered), and \emph{unrendered fabrication} (fabricated and unrendered):
\begin{equation}\label{eq:rates}
\mathrm{transferred}_i=\frac{1}{|\mathcal{D}(i)|}\sum_{f\in\mathcal{D}(i)} a_f\,b_f,
\end{equation}
with $\mathrm{specious}_i$ and $\mathrm{ufab}_i$ from the substitutions
$a_f\!\to\!(1-a_f)$ and, for $\mathrm{ufab}_i$, $b_f\!\to\!(1-b_f)$; the fourth
cell $\frac{1}{|\mathcal{D}(i)|}\sum_{\mathcal{D}(i)} a_f(1-b_f)$ is the
excluded generator loss. All rates average over images.
The headline score is the count form of the transferred rate,
\begin{equation}\label{eq:tf}
\mathrm{FC_{trans}}=\frac{1}{N}\sum_{i=1}^{N}|\mathcal{D}(i)|\,\mathrm{transferred}_i .
\end{equation}
Reconstruction-only cycle
metrics~\citep{ge2024visualfactchecker,cui2025camscore,xu2026itiscore} instead
score the rendered marginal $\frac{1}{N}\sum_{i}\sum_{f\in\mathcal{F}(c_i)}
b_f$---transferred plus specious, up to the unverifiable masses---crediting
fabricated-but-rendered claims as
evidence; we therefore never report it as a caption score and use DINOv3
source--reconstruction similarity as the judge-independent global
check~\citep{simeoni2025dinov3}.
Fact-level scoring makes verdicts attributable (each unsupported
claim names the failing entity, attribute, relation, count, small object, or
action) and
debits generator failures (true but unrendered) to the generator, not the
caption; FactCycle thus operationalizes the recoverability term of $U$ in
\eqref{eq:budgeted_captioning}, instantiated in Section~\ref{sec:eval_protocol}.


\section{Experiments}\label{sec:eval_protocol}

We evaluate the complete evidence path: whether ZoomCap captions preserve
more recoverable, source-grounded facts, which routed components produce the
gain, and whether they improve later VLM training; unless otherwise stated
captions use a Qwen3.5-35B-A3B teacher~\citep{qwen2026qwen35} (protocols in
Appendix~\ref{sec:protocol_details}).

\begin{figure}[!ht]
\centering
\begin{minipage}[b]{0.405\textwidth}
\centering
\vspace{0pt}
\includegraphics[width=\textwidth,height=2.5cm,keepaspectratio]{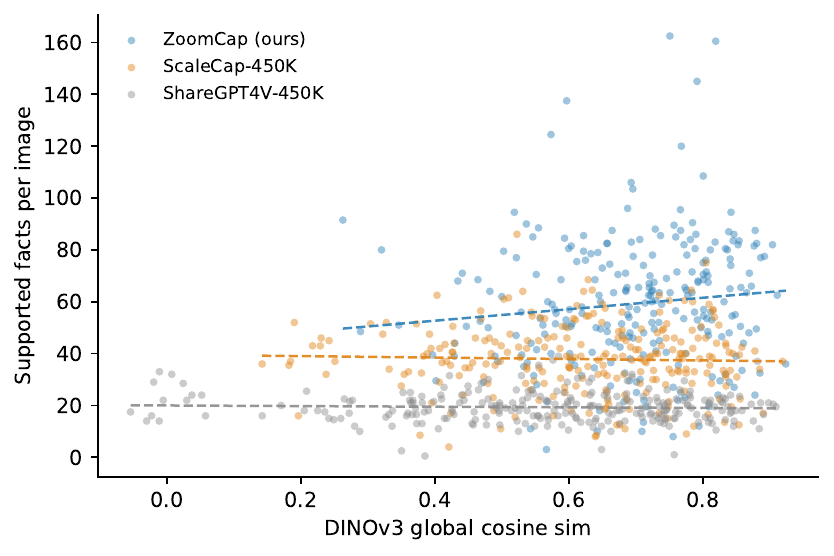}
\subcaption{Per-image rendered claims---a correlate, not a caption score---vs.\ DINOv3 similarity.}
\label{fig:t2i_scatter}
\end{minipage}
\hfill
\begin{minipage}[b]{0.44\textwidth}
\centering
\vspace{0pt}
\includegraphics[width=\textwidth,height=3.0cm,keepaspectratio]{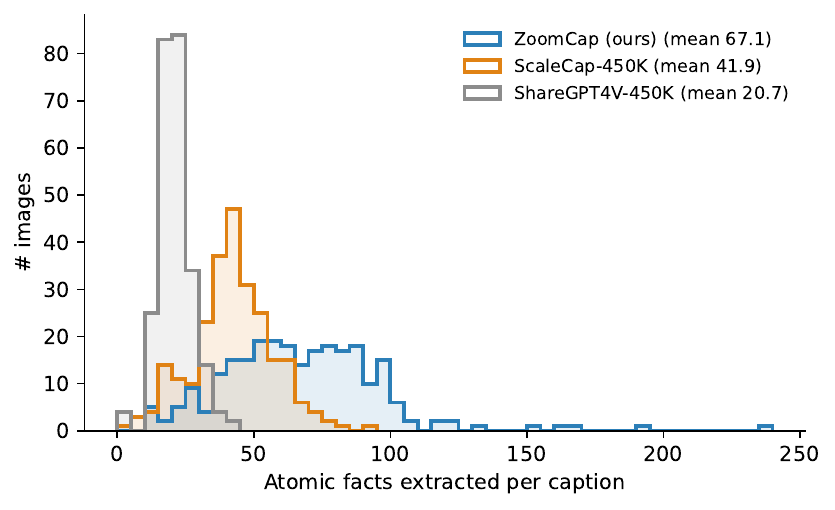}
\subcaption{Atomic facts extracted per caption.}
\label{fig:t2i_fact_density}
\end{minipage}
\caption{\textbf{Information density and generative recoverability in FactCycle.} (a)~each point is one FactCycle-250 image with per-source linear fits; ZoomCap occupies the high-support, high-similarity corner. (b)~mean extracted claims per caption: 67.1 vs.\ 41.9 and 20.7.}
\label{fig:info_density}
\end{figure}


\begin{table}[t]
\centering
\scriptsize
\caption{\textbf{FactCycle master table on FactCycle-250 (two anchors, one blind claim set).}
The same blind claim set is scored on both anchors by the same blind judge
pair (\texttt{Kimi-K3}, \texttt{Qwen3.8-Max}) under one
supported/unsupported/unverifiable rubric; neither judge sees the caption, its
corpus identity, or the other anchor's image (protocol in
Appendix~\ref{sec:dual_anchor_details}). \#Facts/img counts
blind-extracted claims; $\tau$ and $\mathrm{FC_{trans}}$ follow
\eqref{eq:anchors}--\eqref{eq:tf}; the transferred/specious/unrendered-fabrication
shares, with the excluded true-but-unrendered generator cell, partition the
claims decidable on \emph{both} anchors. $\rho$ is prose-only
(Section~\ref{sec:t2i_reconstruction}). DINO is
DINOv3~\citep{simeoni2025dinov3} source--reconstruction similarity, a
judge-independent global check.}
\label{tab:dual_anchor}

\resizebox{0.9\textwidth}{!}{%
\begin{tabular}{lccccccc}
\toprule
\textbf{Caption source} & \textbf{\#Facts/img $\uparrow$} & \textbf{$\tau$: Truthfulness (\%) $\uparrow$} & \textbf{transferred (\%) $\uparrow$} & \textbf{specious (\%) $\downarrow$} & \textbf{\makecell{unrendered\\fab. (\%) $\downarrow$}} & \textbf{$\mathrm{FC_{trans}}$ $\uparrow$} & \textbf{DINO $\uparrow$} \\
\midrule
\midrule
ShareGPT4V-450K~\citep{chen2023sharegpt4v} & 20.7 & 81.4 & 80.3 & 16.6 & 2.0 & 14.8 & 0.566 \\
ScaleCap-450K~\citep{zhang2025scalecap} & 41.9 & 88.9 & 88.2 & 7.4 & 2.6 & 31.5 & 0.620 \\
\rowcolor{LightGray}
\textbf{ZoomCap-450K (Ours)} & \textbf{67.1} & \textbf{93.8} & \textbf{90.9} & \textbf{3.4} & \textbf{1.7} & \textbf{52.4} & \textbf{0.696} \\
\bottomrule
\end{tabular}%
}
\end{table}


\begin{figure}[t]
\centering
\begin{minipage}[b]{0.356\textwidth}
\centering
\vspace{0pt}
\resizebox{\linewidth}{!}{%
\begin{tikzpicture}[scale=0.66, every node/.style={font=\tiny}]
    \def\R{3.0}
    \foreach \r/\lab in {1.5/50,2.1/70,2.7/90,3.0/100} {
        \draw[gray!25] ({\r*cos(90)},{\r*sin(90)}) --
            ({\r*cos(30)},{\r*sin(30)}) --
            ({\r*cos(-30)},{\r*sin(-30)}) --
            ({\r*cos(-90)},{\r*sin(-90)}) --
            ({\r*cos(-150)},{\r*sin(-150)}) --
            ({\r*cos(150)},{\r*sin(150)}) -- cycle;
        \node[gray!60, anchor=west] at (0.06,\r) {\lab};
    }
    \foreach \angle in {90,30,-30,-90,-150,150} {
        \draw[gray!35] (0,0) -- ({\R*cos(\angle)},{\R*sin(\angle)});
    }
    \path[draw=myBlueBase, fill=myBlueBase!22, line width=1.2pt]
        (0,1.893) --
        (1.855,1.071) --
        (1.855,-1.071) --
        (0,-2.079) --
        (-2.245,-1.296) --
        (-1.650,0.953) -- cycle;
    \foreach \x/\y/\v in {
        0/1.893/63.1,
        1.855/1.071/71.4,
        1.855/-1.071/71.4,
        0/-2.079/69.3,
        -2.245/-1.296/86.4,
        -1.650/0.953/63.5
    } {
        \fill[myBlueBase] (\x,\y) circle (1.6pt);
        \node[fill=white, inner sep=1pt] at (\x,\y) {\v};
    }
    \node[align=center] at (0,3.42) {Instruction\\Following};
    \node[align=center] at (3.35,1.9) {Visual\\Accuracy};
    \node[align=center] at (3.35,-1.9) {Information\\Coverage};
    \node[align=center] at (0,-3.45) {Visual\\Reasoning};
    \node[align=center] at (-3.35,-1.9) {Knowledge\\Richness};
    \node[align=center] at (-3.35,1.9) {Expression\\Quality};
    \node[draw=myBlueBase, fill=myBlueBase!22, rounded corners=1pt, inner sep=3pt] at (0,-4.05) {ZoomCap preference score (\%)};
\end{tikzpicture}%
}
\subcaption{Human panel over six dimensions and six annotators.}
\label{fig:human_eval_radar}
\end{minipage}
\hfill
\begin{minipage}[b]{0.57\textwidth}
\centering
\vspace{0pt}
\resizebox{\linewidth}{!}{%
\begin{tikzpicture}
\definecolor{judgeHuman}{HTML}{D64541}
\begin{axis}[
    ybar=1.2pt,
    bar width=2.9pt,
    area legend,
    width=\linewidth,
    height=3.9cm,
    ymin=40, ymax=100,
    xmin=0.3, xmax=8.7,
    xtick={1,...,8},
    xticklabels={Instr.\\follow.,Visual\\accuracy,Info\\coverage,Visual\\reason.,Knowl.\\richness,Expr.\\quality,General,Overall},
    x tick label style={font=\tiny, align=center, yshift=-1pt},
    ylabel={Tie-adjusted preference (\%)},
    ylabel style={font=\scriptsize},
    ytick={40,50,60,70,80,90,100},
    y tick label style={font=\tiny},
    ymajorgrids=false,
    axis lines=left,
    axis line style={gray!60},
    tick align=outside,
    legend columns=2,
    legend style={at={(xticklabel cs:0.5)}, anchor=north, yshift=-3pt, font=\tiny, column sep=6pt, legend cell align=left, draw=none, fill=none},
]
\addplot[fill=judgeHuman, draw=judgeHuman] coordinates {
    (1,63.1) (2,71.4) (3,71.4) (4,69.3) (5,86.4) (6,63.5) (7,81.9) (8,70.9)};
\addplot[fill=myOrange, draw=myOrange] coordinates {
    (1,49.8) (2,63.8) (3,66.7) (4,57.2) (5,91.3) (6,58.4) (7,70.0) (8,77.7)};
\addplot[fill=myGreen, draw=myGreen] coordinates {
    (1,73.3) (2,69.5) (3,80.2) (4,69.7) (5,85.0) (6,81.9) (7,76.2) (8,76.4)};
\addplot[fill=myBlueBase, draw=myBlueBase] coordinates {
    (1,88.1) (2,78.5) (3,92.5) (4,79.5) (5,93.5) (6,84.2) (7,91.7) (8,92.2)};
\draw[dashed, gray] (rel axis cs:0,0.1667) -- (rel axis cs:1,0.1667);
\node[gray!80!black, fill=white, inner sep=1pt, font=\tiny, rotate=90, anchor=west] at (rel axis cs:0.972,0.1667) {chance (50\%)};
\legend{Human, Gemini-3.1-Pro, Qwen3.5-397B-A17B, Kimi-K2.6}
\end{axis}
\end{tikzpicture}%
}
\subcaption{All evaluators; the dashed line marks 50\%.}
\label{fig:judge_eval_bars}
\end{minipage}
\caption{\textbf{Paired-preference evaluation against Qwen3.5-397B-A17B on PrefBench-456.} ZoomCap's largest gains are in knowledge richness and information coverage.}
\label{fig:human_eval_radar_full}
\end{figure}
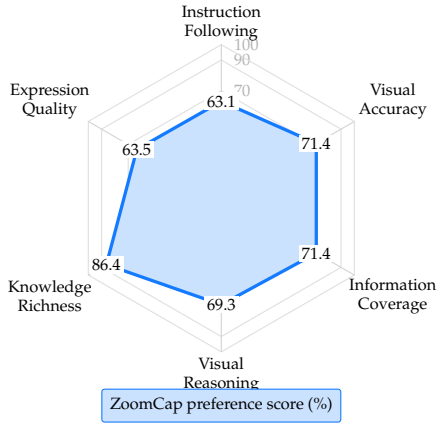
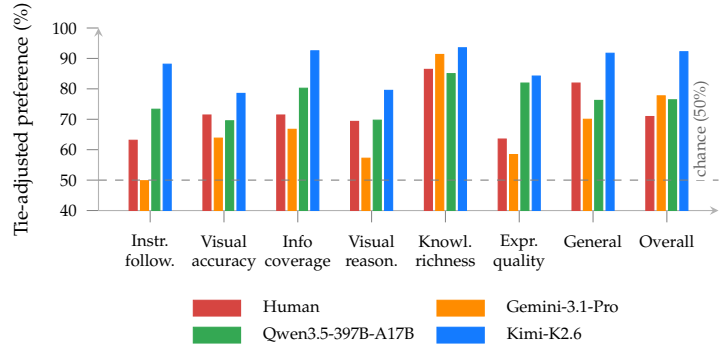


\begin{table}[tb]
\centering
\scriptsize
\caption{\textbf{Evidence-stream attribution (a) and zoom-budget policy (b)
ablations} on PrefBench-456; cells are tie-adjusted preference scores (\%)
(wins plus half ties; sides sum to 100). (a)~judges each configuration against
the same Qwen3.5-397B-A17B single-pass caption (first-level crops only);
(b)~fixes the streams and pairs the two zoom policies directly (rows = two
sides of one head-to-head; crops include the second level).}
\label{tab:ablation}
\resizebox{0.83\textwidth}{!}{%
\begin{tabular}{lccccc}
\toprule
\textbf{Configuration} &
\makecell{\textbf{Final}\\\textbf{score}} &
\makecell{\textbf{Visual}\\\textbf{accuracy}} &
\makecell{\textbf{Information}\\\textbf{coverage}} &
\makecell{\textbf{Knowledge}\\\textbf{richness}} &
\makecell{\textbf{Avg.}\\\textbf{local crops}} \\
\midrule
\multicolumn{6}{l}{\textit{(a) Evidence streams --- opponent: Qwen3.5-397B-A17B single-pass caption}} \\
\midrule
Global draft only (no routed evidence) & 55.9 & 57.9 & 41.6 & 47.6 & 0.0 \\
Agentic evidence w/o retrieved knowledge & 53.0 & 53.0 & 62.4 & 46.7 & 5.0 \\
\rowcolor{LightGray}
Agentic evidence w/ retrieved knowledge & \textbf{74.2} & \textbf{58.0} & \textbf{69.8} & \textbf{92.3} & 5.0 \\
\midrule
\multicolumn{6}{l}{\textit{(b) Zoom budget --- direct head-to-head; the two rows are its two sides}} \\
\midrule
Full second-level pyramid & 28.8 & 36.3 & 23.4 & 50.1 & 10.13 \\
\rowcolor{LightGray}
\textbf{Complexity-gated pyramid (Ours, default)} & \textbf{71.2} & \textbf{63.7} & \textbf{76.6} & 49.9 & \textbf{8.08} \\
\bottomrule
\end{tabular}%
}
\end{table}


\begin{table}[tb]
\centering
\scriptsize
\captionsetup{type=table}
\caption{\textbf{General-domain production adoption (Ling-Tiny-VL-v3).}
Matched 60B-token runs share checkpoint, hyperparameters, and seed; the
experimental arm swaps the production \texttt{vl\_caption} snapshot
(\texttt{pretrain\_caption\_20260623}) for \texttt{zoomcap\_pretrain\_1600w}
(16M ZoomCap captions, up-sampled $4\times$ to a 20\% caption-group share);
sub-items average the four common 40--55B checkpoints.}
\label{tab:adoption_general}
\setlength{\tabcolsep}{4.5pt}
\resizebox{0.83\textwidth}{!}{%
\begin{tabular}{@{}lcc@{\hskip 1.6em}lcc@{}}
\toprule
\multicolumn{3}{@{}l}{\textit{Doc \& OCR suite}} & \multicolumn{3}{l}{\textit{Perception suite}} \\
\textbf{Bench.} & \textbf{Base} & \textbf{+\,ZoomCap} & \textbf{Bench.} & \textbf{Base} & \textbf{+\,ZoomCap} \\
\midrule
ChartQA~\citep{masry2022chartqa}            & 83.7 & \textbf{83.9}\,(+0.2)  & AI2D~\citep{kembavi2016diagram}                & \textbf{82.1} & 81.8 \,(-0.2) \\
DocVQA~\citep{mathew2021docvqa}             & 95.4 & \textbf{95.6}\,(+0.1)  & eyes\_general                                  & \textbf{0.86} & 0.83 \,(-0.03) \\
InfoVQA~\citep{mathew2022infographicvqa}    & 80.2 & \textbf{80.4}\,(+0.2)  & eyes\_vertical                                 & 71.1 & \textbf{71.2}\,(+0.1) \\
OCRBench~\citep{liu2024ocrbench}            & 86.4 & \textbf{86.7}\,(+0.3)  & hallusion\_intern~\citep{guan2024hallusionbench} & 58.6 & \textbf{60.0}\,(+1.4) \\
OCRBenchV2-a~\citep{fu2025ocrbenchv2}       & 40.1 & \textbf{40.1}\,(+0.1)  & MMBench~\citep{liu2024mmbench}                 & 78.6 & \textbf{79.1}\,(+0.5) \\
OCRBenchV2-b~\citep{fu2025ocrbenchv2}       & 46.2 & \textbf{48.5}\,(+2.2)  & MMBench-V11~\citep{liu2024mmbench}             & 84.5 & \textbf{84.9}\,(+0.4) \\
OmniDocBench~\citep{ouyang2025omnidocbench} & 78.0 & \textbf{87.9}\,(+9.8)  & MMStar~\citep{chen2024mmstar}             & \textbf{63.9} & 63.3 \,(-0.6) \\
TableVQA-Bench                              & \textbf{75.8} & 75.1 \,(-0.7)    & MMVet~\citep{yu2024mmvet}                      & 49.9 & \textbf{61.7}\,(+11.8) \\
TextVQA~\citep{singh2019textvqa}            & 80.4 & \textbf{81.0}\,(+0.6)  & RefCOCO~\citep{kazemzadeh2014referitgame,yu2016modeling} & \textbf{87.5} & 87.1 \,(-0.5) \\
\rowcolor{LightGray}
\textbf{Suite mean (9)} & 74.0 & \textbf{75.4}\,(+1.4) & \textbf{Suite mean (9)} & 64.1 & \textbf{65.6}\,(+1.5) \\
\midrule
\multicolumn{3}{@{}l}{\textit{Knowledge VQA suite}} & & \\
SimpleVQA~\citep{cheng2025simplevqa} & 52.4 & \textbf{52.9}\,(+0.5) & & \\
worldvqa\_bench                      & 34.0 & \textbf{35.0}\,(+1.0) & & \\
\cellcolor{LightGray}\textbf{Suite mean (2)} & \cellcolor{LightGray}43.2 & \cellcolor{LightGray}\textbf{43.9}\,(+0.7) & & \\
\bottomrule
\end{tabular}%
}
\end{table}


\begin{table}[tb]
\centering
\scriptsize
\caption{\textbf{Medical-domain production adoption (Ling-Tiny-VL-v3).}
Gemini-3.1-Pro and ZoomCap caption the same 2.64M images under one 64-H800,
$\sim$50B-token recipe (the 128-H800 reference is not compute-matched); Avg.\
gain covers the five medical VQAs.}
\label{tab:adoption_medical}
\resizebox{0.83\textwidth}{!}{%
\begin{tabular}{l c c @{} cccccc}
\toprule
\makecell[l]{\textbf{Medical pretraining recipe}\\\vphantom{\textbf{GMAI}}} &
\makecell{\textbf{Medical}\\\textbf{images}} &
\makecell{\textbf{Hallusion}\\\textbf{(intern)}} &
\makecell{\textbf{GMAI}\\\textbf{MMBench}} &
\makecell{\textbf{PathVQA}\\\vphantom{\textbf{GMAI}}} &
\makecell{\textbf{PMC}\\\textbf{VQA}} &
\makecell{\textbf{SLAKE}\\\vphantom{\textbf{GMAI}}} &
\makecell{\textbf{VQA}\\\textbf{RAD}} &
\makecell{\textbf{Avg.}\\\textbf{gain}} \\
\midrule
Reference recipe (128$\times$H800) & 3.66M & \textbf{60.9} & 55.3 & 39.4 & 60.9 & 68.6 & 64.3 & -- \\
+\,Gemini-3.1-Pro captions (64$\times$H800) & +2.64M & 59.4 & 59.3 & \textbf{40.0} & 62.8 & \textbf{71.3} & 63.9 & +1.76 \\
\rowcolor{LightGray}
+\,\textbf{ZoomCap captions (Ours; 64$\times$H800)} & +2.64M & 60.4 & \textbf{61.4} & 39.7 & \textbf{64.2} & 71.0 & \textbf{67.4} & \textbf{+3.04} \\
\bottomrule
\end{tabular}%
}
\end{table}

\begin{table}[tb]
\centering
\scriptsize
\caption{\textbf{CPT comparison with different 450K caption datasets on 11 benchmarks.} Within each backbone block only the CPT caption corpus varies; all methods are re-evaluated with Qwen3.5-397B-A17B for judge-dependent benchmarks.}
\label{tab:cpt_comparison}
\setlength{\tabcolsep}{2.05pt}
\renewcommand{\arraystretch}{0.85}
\resizebox{0.93\textwidth}{!}{%
\begin{tabular}{@{}llcccccccccccc@{}}
\toprule
\textbf{Model} & \makecell{\textbf{Pretraining}\\\textbf{Data}} &
\makecell{\textbf{Info}\\\textbf{VQA}} &
\makecell{\textbf{Doc}\\\textbf{VQA}} &
\makecell{\textbf{Chart}\\\textbf{QA}} &
\makecell{\textbf{MM}\\\textbf{Star}} &
\makecell{\textbf{Math}\\\textbf{Vista}} &
\makecell{\textbf{LLaVA}\\\textbf{Bench}} &
\textbf{MMVet} & \textbf{MMB} & \textbf{MMMU} & \textbf{SEED} & \textbf{AI2D} & \textbf{Avg.} \\
\midrule
\multirow{5}{*}{\makecell[l]{Qwen2.5-7B \\+ Qwen2.5-ViT}}
& Vanilla & 43.66 & 77.25 & 73.56 & 45.93 & 47.20 & 51.70 & 48.35 & 72.47 & 44.78 & 71.80 & 74.03 & 59.16 \\
& ShareGPT4V-450K & 45.80 & 79.46 & 74.20 & 48.87 & \textbf{49.60} & 44.70 & 49.22 & 72.52 & 42.89 & 72.38 & 74.38 & 59.46 \\
& DenseFusion-450K & 47.94 & 81.48 & 74.36 & 50.80 & 45.20 & 54.00 & 48.17 & 72.05 & 45.00 & 71.66 & 74.16 & 60.44 \\
& ScaleCap-450K & 49.02 & 81.94 & \textbf{74.96} & 50.39 & 46.60 & 50.00 & 47.39 & 73.71 & 43.29 & 72.32 & 75.68 & 60.48 \\
\rowcolor{LightGray}
& \textbf{ZoomCap-450K (Ours)} & \textbf{51.08} & \textbf{82.64} & 74.76 & \textbf{52.66} & 49.30 & \textbf{54.80} & \textbf{52.75} & \textbf{74.65} & \textbf{48.40} & \textbf{72.72} & \textbf{75.94} & \textbf{62.70} \\
\midrule
\multirow{5}{*}{\makecell[l]{Qwen2.5-3B \\+ Qwen2.5-ViT}}
& Vanilla & 37.35 & 70.11 & 70.24 & 47.13 & 44.70 & 43.80 & 42.20 & 68.36 & 39.00 & 69.56 & 70.76 & 54.84 \\
& ShareGPT4V-450K & 40.85 & 74.13 & 71.88 & 48.73 & 44.50 & 48.80 & 44.86 & 69.40 & 39.67 & 70.45 & 72.64 & 56.90 \\
& DenseFusion-450K & 45.16 & 77.78 & 71.88 & 48.33 & 46.40 & 52.20 & 46.61 & 70.81 & 39.33 & 70.90 & 73.19 & 58.42 \\
& ScaleCap-450K & 45.63 & 80.04 & 73.04 & 44.13 & 45.80 & 50.70 & 49.27 & 71.53 & 41.22 & 70.05 & 69.01 & 58.22 \\
\rowcolor{LightGray}
& \textbf{ZoomCap-450K (Ours)} & \textbf{49.71} & \textbf{81.43} & \textbf{74.24} & \textbf{48.93} & \textbf{46.90} & \textbf{55.80} & \textbf{51.83} & \textbf{72.05} & \textbf{41.78} & \textbf{71.90} & \textbf{74.26} & \textbf{60.80} \\
\midrule
\multirow{5}{*}{\makecell[l]{Ling-mini-base-2.0 \\ + Qwen2.5-ViT}}
& Vanilla & 38.39 & 69.38 & 70.96 & 46.40 & 47.30 & 43.10 & 39.82 & 69.82 & 42.78 & 69.62 & 72.86 & 55.49 \\
& ShareGPT4V-450K & 42.71 & 72.67 & 72.52 & 47.53 & 47.20 & 45.60 & 49.13 & 70.58 & 43.11 & 70.48 & 73.42 & 57.72 \\
& DenseFusion-450K & 46.47 & 76.14 & 74.16 & 48.00 & 47.90 & \textbf{48.90} & 49.50 & 69.24 & 43.04 & 71.61 & 72.90 & 58.90 \\
& ScaleCap-450K & 46.55 & 75.50 & 74.96 & 47.00 & 49.70 & 36.60 & 46.83 & 70.70 & \textbf{44.33} & 71.54 & \textbf{74.64} & 58.03 \\
\rowcolor{LightGray}
& \textbf{ZoomCap-450K (Ours)} & \textbf{48.22} & \textbf{77.42} & \textbf{75.56} & \textbf{49.93} & \textbf{50.80} & 47.40 & \textbf{50.14} & \textbf{72.10} & 44.22 & \textbf{71.95} & 74.55 & \textbf{60.21} \\
\bottomrule
\end{tabular}%
}
\end{table}

\subsection{Direct Caption Quality}

\subsubsection{FactCycle: Fact-Level Truth and Realizability}\label{sec:t2i_reconstruction}

\begin{figure}[t]
\centering
\includegraphics[width=0.92\textwidth]{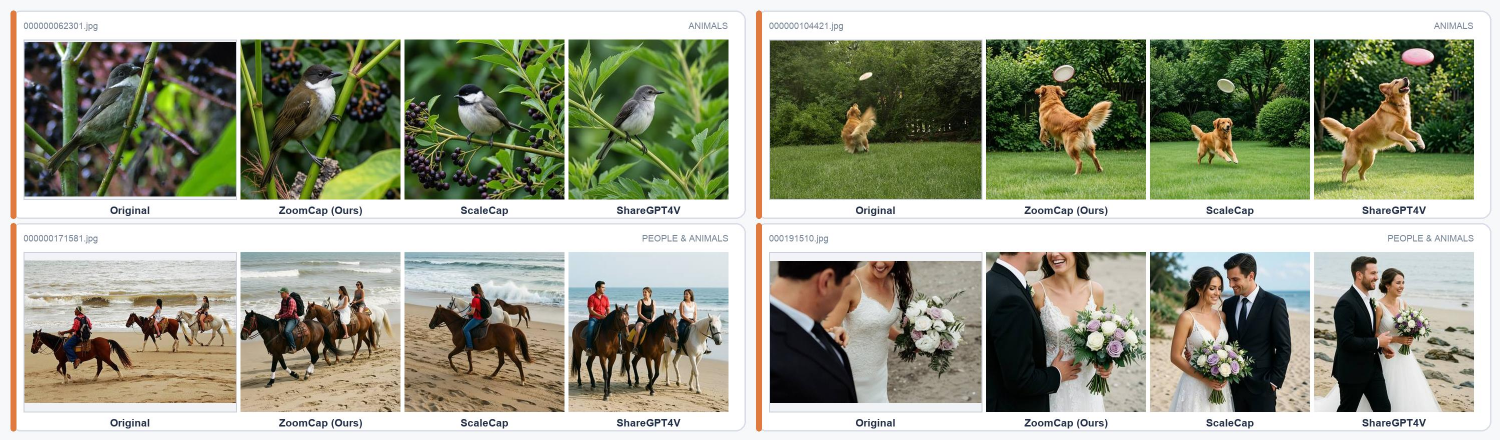}
\vskip -1ex
\caption{\textbf{Qualitative FactCycle reconstruction comparison.} Each row uses captions of the same source image to condition the fixed text-to-image generator; ZoomCap reconstructions preserve more of the source composition and fine-grained content; the remaining fourteen FactCycle-250 pairs appear in Appendix Figure~\ref{fig:t2i_more_vis}.}
\label{fig:t2i_reconstruction}
\end{figure}

FactCycle (Section~\ref{sec:factcycle_metric}) is instantiated on
\textbf{FactCycle-250}, a paired evaluation pool of 250 source images shared
by ZoomCap-450K, ScaleCap-450K~\citep{zhang2025scalecap}, and
ShareGPT4V-450K~\citep{chen2023sharegpt4v}, balanced across five scene
types, all rendered by one fixed
\texttt{Qwen-Image-2512-Lightning} generator~\citep{wu2025qwenimage}
(only the caption varies; Figure~\ref{fig:t2i_reconstruction} previews
reconstructions).
ZoomCap attains Truthfulness $\tau=93.8\%$ versus $88.9\%$ and $81.4\%$
(paired $p=4.57\times10^{-21}$ and $1.69\times10^{-21}$), and Table~\ref{tab:dual_anchor} shows where the anchors diverge: ZoomCap keeps 90.9\% of its decidable claims transferred and admits
only 3.4\% specious ones against $16.6\%$ for ShareGPT4V; aggregating the
transferred quadrant, ZoomCap realizes 52.4 transferred facts per image
($\mathrm{FC_{trans}}$), $1.7\times$ ScaleCap and $3.5\times$ ShareGPT4V.
Realizability meanwhile \emph{reverses} ($\rho=92.3$/$93.6$/$95.7$): captions
that state less render more completely, so single-anchor realizability ranks
these captions backwards; the two anchors disentangle this by charging
true-but-unrendered losses to the generator, not the caption
(Figure~\ref{fig:factcycle_dual_anchor}).
The higher claim count (67.1 per image versus 41.9 and 20.7) is corroborated
by judge-independent DINOv3~\citep{simeoni2025dinov3} similarity (0.696 versus
0.620 and 0.566; Figures~\ref{fig:t2i_scatter}--\ref{fig:t2i_fact_density});
counts scale with length but rates do not---ZoomCap is $1.45\times$ longer
than ScaleCap yet leads every rate column of Table~\ref{tab:dual_anchor}.
The gain is not confined to one slice: the $\mathrm{FC_{trans}}$ ordering
holds in every scene bucket (Table~\ref{tab:dual_anchor_scenes}), and at
category level ZoomCap keeps the lowest specious share in all seven
categories and the highest transferred share in six, small objects being the
single exception (Table~\ref{tab:dual_anchor_categories}); breakdowns,
conservative accounting, and judge robustness are in
Appendix~\ref{sec:dual_anchor_details}.

\subsubsection{Source-Grounded Preference}\label{sec:human_eval}

We compare ZoomCap with Qwen3.5-397B-A17B on \textbf{PrefBench-456}, a
human-annotated paired-preference benchmark of 456 images fixed-seed sampled
from an internal business-image pool (anonymized pairs, randomized order). Six annotators and three VLM judges
(Gemini-3.1-Pro~\citep{google2026gemini31pro};
Qwen3.5-397B-A17B~\citep{qwen2026qwen35}, which supplies the opposing
captions; Kimi-K2.6~\citep{moonshot2026kimik26}) score each pair on six axes
(Table~\ref{tab:judge_eval}); we report tie-adjusted preference scores: the
human general-perception verdict and each judge's final overall verdict.
Under this convention humans give ZoomCap 81.9\% and every VLM judge agrees
(76.4--92.2\%); knowledge richness is strongest for all evaluators
(85.0--93.5\%) and coverage improves consistently
(Figure~\ref{fig:human_eval_radar_full}, Table~\ref{tab:judge_eval},
Appendix Figure~\ref{fig:qualitative_cases}).

\subsection{What Produces the Gain?}\label{sec:multi_judge}

Table~\ref{tab:ablation}(a), against the single-pass opponent: the draft is
visually accurate (57.9) but misses coverage (41.6); local observations lift
coverage to 62.4 without stabilizing the verdict (53.0); retrieved knowledge
then raises knowledge richness from 46.7 to 92.3 and the score to 74.2.
Panel (b) varies only the zoom policy: in a direct head-to-head,
complexity-gated zooming takes 71.2 of the score to the unrestricted pyramid's
28.8 with 20.2\% fewer local and 39.8\% fewer second-level crops---gating
removes unverified observations, not evidence.

\subsection{Transfer to VLM Pre-Training}\label{sec:adoption}

The general-domain pair shares checkpoint, hyperparameters, seed, and a
60B-token budget; the medical runs caption the same 2.64M images under a shared 64-GPU, $\sim$50B-token recipe (reference:
128 GPUs, $\sim$100B tokens).
In production, ZoomCap captions improve document/OCR by 1.4 points and
perception by 1.5 points
(Table~\ref{tab:adoption_general}; Appendix Figure~\ref{fig:adoption_curves})
and the medical five-VQA mean by 3.04 over the reference and 1.28 over
same-image Gemini-3.1-Pro captions (Table~\ref{tab:adoption_medical},
Appendix~\ref{sec:medical_adoption}); in controlled CPT they beat the strongest
competing 450K corpus by 1.31--2.38 points on the 11-benchmark mean across all
three backbones (Table~\ref{tab:cpt_comparison}), with matching gains in
caption scale and entity density (Table~\ref{tab:dataset_stats}).


\section{Conclusion}

We formulated fine-grained captioning as recoverable visual information
transfer and presented ZoomCap, an evidence-routed teacher that conditions on a
global draft to acquire knowledge, OCR, and zoom-in evidence with
geometry-grounded relations, then fuses only novel, consistent, non-duplicate
evidence. To measure more than verbosity we introduced FactCycle, a two-anchor protocol
scoring a caption's atomic claims against the source image (truthfulness) and a
fixed reconstruction (realizability), separating fabrication from generator
loss claim by claim. On FactCycle-250 and PrefBench-456, with controlled pre-training and production
adoption, these captions preserve more recoverable content than existing
datasets and strong single-pass models and improve vision-language training
across backbones and domains.

\subsection*{AI use statement}

Generative AI tools were used solely to aid and polish the writing of this
manuscript, including drafting, revising, translating, and formatting text to
meet publication guidelines. No AI tools were used to generate experimental
data, conduct model calls, or produce scientific figures; all results and
visuals are original. The authors manually verified all AI-assisted content and
take full responsibility for the final manuscript.

\subsection*{Ethics statement}

We report only aggregate annotation results and no annotator identifiers; the
evaluation-benchmark images, captions, and intermediate artifacts are not
released, and any future release remains subject to the applicable privacy and
licensing review. The medical-domain results are benchmark evaluations of pre-training
recipes, not clinical validation; ZoomCap captions and the resulting models must
not be used to make diagnostic or treatment decisions without domain-specific
validation and oversight.

\subsection*{Reproducibility statement}

Section~\ref{sec:pipeline} specifies the evidence-routing workflow, and
Appendix~\ref{sec:impl_details} records the implemented complexity and geometry
rules. Section~\ref{sec:factcycle_metric} defines the FactCycle metrics and
Section~\ref{sec:eval_protocol} the comparisons.
Appendix~\ref{sec:protocol_details} records the FactCycle sampling,
models, seeds, aggregation, human-evaluation protocol, and controlled
continual-pre-training recipe; the multi-judge and corpus-statistics breakdowns
appear in Appendices~\ref{sec:judge_eval_details}
and~\ref{sec:corpus_stats}. The controlled training
comparisons use one fixed seed on the publicly released ZoomCap-450K corpus
(page-1 footnote), while the internal production datasets cannot be
redistributed; we state these constraints rather than claiming full public
reproducibility.

\clearpage

\bibliographystyle{antgroup}

\bibliography{ref/reference}

\clearpage
\appendix

\section{Evaluation and Training Protocol Details}
\label{sec:protocol_details}

\paragraph{Teacher and recorded evidence.}
ZoomCap uses Qwen3.5-35B-A3B~\citep{qwen2026qwen35} as the teacher VLM and \texttt{PaddleOCR-VL-1.5}~\citep{baidu2026paddleocrvl} as the OCR engine unless
otherwise stated. The workflow in Figure~\ref{fig:pipeline} records its draft,
search context, OCR summary, grounded boxes, patch coordinates and captions,
attribute-augmented subject--predicate--object--environment (A-SPOE) facts, routing decisions, latency, and token use. These artifacts make
the evidence stream auditable and modules individually ablatable.

\paragraph{Three evaluation levels.}
At the caption level, FactCycle measures claim-level truthfulness and
generative recoverability on two blind anchors while human and multi-judge
comparisons inspect captions against the source
image; at the mechanism level, evidence-stream and zoom-budget ablations isolate
why the workflow improves; at the data level, controlled 450K continual
pre-training (CPT) changes only the caption corpus, and two production studies
test transfer to general and medical pre-training mixtures.

\paragraph{From teacher workflow to reusable supervision.}
The workflow output can be used directly as a high-fidelity caption, but its
larger value in this paper is as reusable supervision: given images
$\{I_j\}_{j=1}^{M}$, ZoomCap produces caption pairs $\{(I_j,y_j)\}$ for
continual pre-training. The controlled ZoomCap-450K comparison and production
adoption studies (Section~\ref{sec:adoption}) therefore test whether the
evidence acquired by the teacher remains useful after it is converted into
training data.

\paragraph{FactCycle sampling and aggregation.}
We intersect the image pools underlying ZoomCap-450K, ScaleCap-450K~\citep{zhang2025scalecap}, and
ShareGPT4V-450K~\citep{chen2023sharegpt4v}, remove exact duplicate rows, shuffle with seed 42, and select the
first 50 accepted examples in each of five categories: indoor, outdoor nature,
urban/street, food, and people/animals, forming \textbf{FactCycle-250}. A text-only scene classifier sees only the
ScaleCap caption during stratification. Screenshots, documents, charts, collages,
artwork, and illustrations are excluded, so this experiment does not test the OCR
branch. Every caption is reconstructed at $1024\!\times\!1024$ with
\texttt{Qwen-Image-2512-Lightning}~\citep{wu2025qwenimage} and seed 42; the
reconstruction prompt is truncated at 3{,}000 characters for every caption
source, because in engineering sweeps of this generator renderable detail peaks
near 2.8k prompt characters and falls off when less of the caption reaches it.
The truncation
binds asymmetrically---on 108 of 250 ZoomCap images, 14 of 250 ScaleCap
images, and no ShareGPT4V image---so $\rho$ and $\mathrm{FC_{trans}}$ are
conservative for ZoomCap: restricted to the 142 images whose ZoomCap prompt was
not truncated, the ordering and both significance tests are unchanged
($\mathrm{FC_{trans}}$ $49.4/30.7/14.3$; paired $p=1.0\times10^{-8}$ and
$1.2\times10^{-12}$), and the truncated images still transfer more claims per
image than the untruncated ones ($56.3$ vs.\ $49.4$). A text-only
\texttt{Kimi-K3}~\citep{moonshot2026kimik3} extractor produces the atomic-claim set; independent
\texttt{Kimi-K3} and \texttt{Qwen3.8-Max}~\citep{qwen2026qwen38} judges then see only the
reconstruction and anonymized claims.
The truthfulness anchor reuses the identical claim set, with the same
\texttt{Kimi-K3} and \texttt{Qwen3.8-Max} judges seeing only the source
image and the anonymized claims
(Section~\ref{sec:dual_anchor_details}). Metrics are computed per image by first
averaging the two judges, then averaging over FactCycle-250. Across the three caption
sources, the reconstruction anchor yields exact inter-judge agreement of
0.933 over the 32{,}205 claims both realizability judges verdicted
(64{,}617 judgments).

\paragraph{Paired preference.}
The human study uses a two-alternative forced-choice protocol with randomized
presentation order on \textbf{PrefBench-456}, our benchmark of 456
image--caption pairs fixed-seed sampled from an internal business-image pool
(one anonymized caption pair per image, human-annotated). Six independent
annotators assess each pair across six dimensions. Each VLM judge evaluates
all 456 pairs with the same rubric. We report tie-adjusted preference scores
(wins plus half the ties, so a pairing's two sides sum to 100), matching
the original ZoomCap evaluation.
The ablations use the same convention, judged by
Gemini-3.1-Pro~\citep{google2026gemini31pro} against the opponents named in
Table~\ref{tab:ablation}; panel (b) pairs the two zoom policies directly, so
its two rows are the two sides of one head-to-head over the same 456 pairs.

\paragraph{Controlled CPT recipe.}
All controlled 450K runs use a three-stage recipe: projector initialization on
BLIP-558K~\citep{zhu2023minigpt4}, one epoch of CPT on the indicated 450K caption corpus, and one epoch of
instruction tuning on Open-LLaVA-NeXT-Instruct-1M~\citep{li2024llavanextinterleave}. During CPT and instruction
tuning the vision tower is frozen while the projector and full language model are
trainable. Images use a 589{,}824-pixel maximum (a $768\!\times\!768$ equivalent).
CPT uses learning rate $4\times10^{-5}$ and effective batch size 256; instruction
tuning uses $2\times10^{-5}$ and effective batch size 128. All runs use seed 42.
Within each backbone block, the initialization, training data outside the CPT
caption corpus, resolution, and evaluation protocol are held fixed.

\section{Routing and Zoom Implementation Details}
\label{sec:impl_details}

\paragraph{Caption-coverage gap recovery.}
Let $\Omega_{\mathcal{E}}=\bigcup_i b_i$ be the union of grounded boxes already
explained by the draft caption. The implementation enumerates rectangles whose
edges come from image and box boundaries and selects the largest rectangle that
does not intersect $\Omega_{\mathcal{E}}$; a second gap can be selected after the
first is added to the covered set. Near-full-image entity boxes and extremely
narrow gaps are rejected. Unlike generic saliency, these crops are defined
relative to what the draft has already grounded.

\paragraph{Complexity-gated zooming.}
Each first-level patch is captioned independently and retained with its source
coordinates. The complexity score $C(p)$
(\eqref{eq:complexity} in the main text) nominates patches for a second
observation. Here $N_{\mathrm{ent}}$ is the number of entity boxes with intersection-over-box
at least 0.25, $N_{\mathrm{small}}$ counts those covering at most 1.5\% of the
patch, $\rho_{\mathrm{ent}}$ is clipped entity-intersection area divided by patch
area, and $r_{\mathrm{area}}$ is the patch-to-image area ratio. We set the
score weights empirically to $w_{\mathrm{ent}}=1.0$, $w_{\mathrm{small}}=0.8$,
$w_{\mathrm{clip}}=2.0$, and $w_{\mathrm{area}}=0.3$. A first-level patch is
eligible when $N_{\mathrm{ent}}\ge6$, $N_{\mathrm{small}}\ge3$, or
$N_{\mathrm{ent}}\ge4$ and $\rho_{\mathrm{ent}}\ge0.35$; an agent judgment then
vetoes nominations whose zoom it rates uninformative. The budgeted policy keeps
broad first-level coverage and passes only the highest-scoring parent approved
by both gates, with its already-generated children, to fusion, which isolates
the fusion-time patch budget from generation randomness.

\paragraph{Attribute-augmented subject--predicate--object--environment (A-SPOE) acceptance rules.}
The implementation instantiates \eqref{eq:aspoe_margin} in the main
text: for each entity pair it normalizes horizontal and vertical
center gaps by the pair's mean box width and height, and rejects pairs with IoU
above 0.5 ($\tau_i$), maximum normalized gap below 0.5 ($\tau_g$), or dominant-to-minor gap ratio
below 1.3 ($\tau_r$). Otherwise it assigns left/right or above/below from the dominant
axis, ranks candidates by the dominant gap times this ratio, removes duplicate
subject--relation edges, caps each object at two incoming relations, and retains
at most eight relations. Thus the reported relations follow explicit box
geometry rather than an unimplemented learned relation score.

\paragraph{Draft-anchored gated fusion.}
At temperature zero the fusion model receives the draft and the ordered patch
descriptions and outputs only a new visual-details section (text, materials,
colors, small objects, counts, local arrangements) that adds nothing already
stated and excludes encyclopedic knowledge; the code appends this section to
the draft, and geometry-derived relations pass through a constrained prompt
that preserves their supplied direction. The novelty, consistency, and overlap
gates of \eqref{eq:fusion_gate} are therefore realized training-free.

\section{Multi-Judge Cross-Validation}
\label{sec:judge_eval_details}

{\centering
\includegraphics[width=\textwidth]{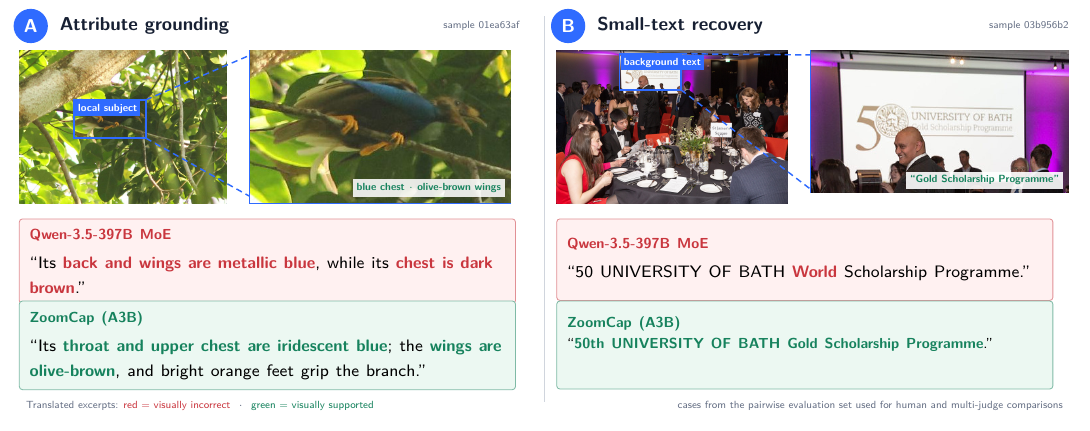}
\captionof{figure}{\textbf{Representative cases from PrefBench-456.} The left case tests localized attribute grounding, the right tests small-text recovery (translated excerpts; red marks visually incorrect spans, green image-supported spans). ZoomCap preserves local evidence that Qwen3.5-397B-A17B reverses or misreads.}
\label{fig:qualitative_cases}
\par}


\noindent\begin{minipage}{\textwidth}
\centering
\scriptsize
\captionsetup{type=table}
\caption{\textbf{Multi-judge cross-validation of ZoomCap vs.\ Qwen3.5-397B-A17B on PrefBench-456.} The human row aggregates six annotators; cells are tie-adjusted preference scores (\%, wins plus half ties, so a pairing's sides sum to 100). General is the general-perception verdict and Overall the final paired-comparison verdict (human row: mean of the six dimension rates). Headline rates in Section~\ref{sec:human_eval} are the human \emph{General} verdict (81.9) and each judge's \emph{Overall} verdict (76.4--92.2). Cell shading marks the rate bins: \colorbox{red!30}{\rule{0.9em}{0pt}\rule{0pt}{0.7ex}}\,$<$50, \colorbox{blue!10}{\rule{0.9em}{0pt}\rule{0pt}{0.7ex}}\,50--60, \colorbox{blue!18}{\rule{0.9em}{0pt}\rule{0pt}{0.7ex}}\,60--70, \colorbox{blue!28}{\rule{0.9em}{0pt}\rule{0pt}{0.7ex}}\,70--80, \colorbox{blue!40}{\rule{0.9em}{0pt}\rule{0pt}{0.7ex}}\,80--90, \colorbox{blue!50}{\rule{0.9em}{0pt}\rule{0pt}{0.7ex}}\,$\geq$90.}
\label{tab:judge_eval}
\resizebox{\textwidth}{!}{%
\begin{tabular}{lcccccccc}
\toprule
\textbf{Judge} & \makecell{\textbf{Instr.}\\\textbf{follow.}} & \makecell{\textbf{Visual}\\\textbf{accuracy}} & \makecell{\textbf{Info}\\\textbf{coverage}} & \makecell{\textbf{Visual}\\\textbf{reasoning}} & \makecell{\textbf{Knowledge}\\\textbf{richness}} & \makecell{\textbf{Expr.}\\\textbf{quality}} & \textbf{General} & \textbf{Overall} \\
\midrule
Human (6 annotators) & \cellcolor{blue!18}63.1 & \cellcolor{blue!28}71.4 & \cellcolor{blue!28}71.4 & \cellcolor{blue!18}69.3 & \cellcolor{blue!40}86.4 & \cellcolor{blue!18}63.5 & \cellcolor{blue!40}81.9 & \cellcolor{blue!28}70.9 \\
Gemini-3.1-Pro~\citep{google2026gemini31pro} & \cellcolor{red!30}49.8 & \cellcolor{blue!18}63.8 & \cellcolor{blue!18}66.7 & \cellcolor{blue!10}57.2 & \cellcolor{blue!50}91.3 & \cellcolor{blue!10}58.4 & \cellcolor{blue!28}70.0 & \cellcolor{blue!28}77.7 \\
Qwen3.5-397B-A17B~\citep{qwen2026qwen35} & \cellcolor{blue!28}73.3 & \cellcolor{blue!18}69.5 & \cellcolor{blue!40}80.2 & \cellcolor{blue!18}69.7 & \cellcolor{blue!40}85.0 & \cellcolor{blue!40}81.9 & \cellcolor{blue!28}76.2 & \cellcolor{blue!28}76.4 \\
Kimi-K2.6~\citep{moonshot2026kimik26} & \cellcolor{blue!40}88.1 & \cellcolor{blue!28}78.5 & \cellcolor{blue!50}92.5 & \cellcolor{blue!28}79.5 & \cellcolor{blue!50}93.5 & \cellcolor{blue!40}84.2 & \cellcolor{blue!50}91.7 & \cellcolor{blue!50}92.2 \\
\bottomrule
\end{tabular}%
}
\end{minipage}

\section{More Qualitative FactCycle Reconstructions}
\label{fig:t2i_more_vis_sec}

{\centering
\includegraphics[width=\textwidth,height=0.62\textheight,keepaspectratio]{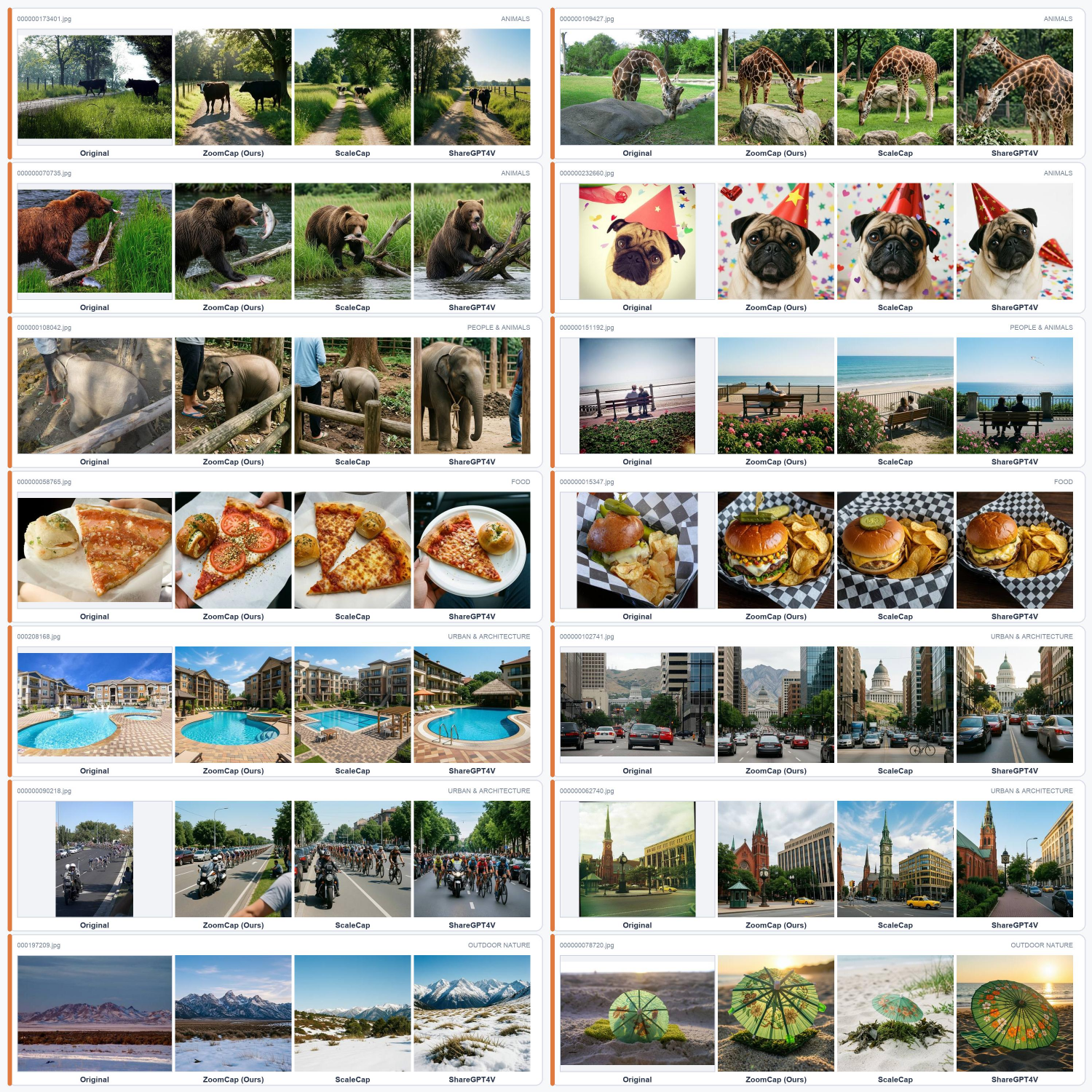}
\captionof{figure}{\textbf{Additional FactCycle reconstruction comparisons.} The fourteen paired sources not shown in Figure~\ref{fig:t2i_reconstruction}, generated with the identical fixed generator and caption sources.}
\label{fig:t2i_more_vis}
\par}

\section{Additional FactCycle Breakdowns}
\label{sec:dual_anchor_details}

On the 250 paired images (five scene buckets, 50 each) this appendix expands
Table~\ref{tab:dual_anchor}: each caption's blind claim set is scored on two
anchors, the source image (Truthfulness $\tau$) and the fixed-generator
reconstruction (Realizability $\rho$).

\paragraph{Aggregation.}
$\tau$ and $\rho$ average the two judges within an image, then average over
images (Appendix~\ref{sec:protocol_details}). The quadrant rates of
\eqref{eq:rates} and $\mathrm{FC_{trans}}$ of \eqref{eq:tf} come from one
cross-tabulation of the same verdicts: truthfulness rows are pooled over
judges, the reconstruction leg contributes its per-claim majority, and claims
on which the two reconstruction judges split are dropped
(2{,}462/1{,}301/444 rows). The denominator is thus the rows decidable on both
anchors (28{,}794/17{,}854/9{,}226), the four cells---transferred, specious,
unrendered fabrication, and the excluded true-but-unrendered generator
cell---sum to $100\%$, and $\mathrm{FC_{trans}}$ divides the both-supported
rows (26{,}184/15{,}749/7{,}411) by $250\times2$. \#Facts/img counts
blind-extracted claims throughout;
Table~\ref{tab:dual_anchor_conservative} instead keeps
reconstruction-unverifiable rows in the unrealized denominator.

\paragraph{Robustness to the aggregation order.}
Averaging judge scores per claim before intersecting the anchors, and counting
a claim as decidable if either judge decided it, moves the quadrant shares by
at most 3.0 points and $\mathrm{FC_{trans}}$ by 4.1 (transferred
88.0/85.3/79.3; specious 4.3/8.2/16.4; unrendered fabrication 2.2/3.5/2.7;
$\mathrm{FC_{trans}}$ 56.5/33.7/15.6), preserving every ordering and both
significance tests.

\paragraph{Breakdowns.}
Every scene bucket preserves the overall $\mathrm{FC_{trans}}$ ordering
(Table~\ref{tab:dual_anchor_scenes}); by category
(Table~\ref{tab:dual_anchor_categories}) the transferred and specious shares
differ in one cell only---small objects, where ScaleCap leads 85.4 vs.\
84.2\%.


\begin{table}[htbp]
\centering
\scriptsize
\caption{\textbf{Two-anchor FactCycle scoring per scene bucket (50 images
each).} Columns follow Table~\ref{tab:dual_anchor}: \#Facts/img counts
blind-extracted claims, Truthfulness $\tau$ is
supported/(supported+unsupported) against the source image; the transferred,
specious, and unrendered-fabrication shares cover claims decidable on both
anchors and, with the excluded true-but-unrendered generator cell, sum to
100\%. Per-bucket realizability is discussed in the Realizability accounting
paragraph below.}
\label{tab:dual_anchor_scenes}
\resizebox{\textwidth}{!}{%
\begin{tabular}{llcccccc}
\toprule
\textbf{Scene} & \textbf{Caption source} & \textbf{\#Facts/img $\uparrow$} & \textbf{$\tau$ (\%) $\uparrow$} & \textbf{transferred (\%) $\uparrow$} & \textbf{specious (\%) $\downarrow$} & \textbf{\makecell{unrendered\\fab. (\%) $\downarrow$}} & \textbf{$\mathrm{FC_{trans}}$ $\uparrow$} \\
\midrule
\midrule
\multirow{3}{*}{Indoor}
 & ShareGPT4V-450K~\citep{chen2023sharegpt4v} & 19.6 & 81.6 & 79.0 & 17.3 & 2.4 & 14.3 \\
 & ScaleCap-450K~\citep{zhang2025scalecap} & 40.2 & 88.6 & 87.1 & 8.3 & 2.6 & 30.3 \\
 & \cellcolor{LightGray}\textbf{ZoomCap-450K (Ours)} & \cellcolor{LightGray}\textbf{68.8} & \cellcolor{LightGray}\textbf{93.2} & \cellcolor{LightGray}\textbf{88.6} & \cellcolor{LightGray}\textbf{4.0} & \cellcolor{LightGray}\textbf{2.0} & \cellcolor{LightGray}\textbf{52.4} \\
\midrule
\multirow{3}{*}{\makecell{Outdoor\\nature}}
 & ShareGPT4V-450K~\citep{chen2023sharegpt4v} & 19.7 & 81.2 & 78.7 & 18.7 & \textbf{1.6} & 14.0 \\
 & ScaleCap-450K~\citep{zhang2025scalecap} & 39.0 & 89.5 & 89.5 & 6.3 & 2.5 & 30.1 \\
 & \cellcolor{LightGray}\textbf{ZoomCap-450K (Ours)} & \cellcolor{LightGray}\textbf{58.8} & \cellcolor{LightGray}\textbf{94.1} & \cellcolor{LightGray}\textbf{91.8} & \cellcolor{LightGray}\textbf{3.6} & \cellcolor{LightGray}\textbf{1.6} & \cellcolor{LightGray}\textbf{46.3} \\
\midrule
\multirow{3}{*}{\makecell{Urban\\street}}
 & ShareGPT4V-450K~\citep{chen2023sharegpt4v} & 22.5 & 80.1 & 79.7 & 16.9 & 2.3 & 15.9 \\
 & ScaleCap-450K~\citep{zhang2025scalecap} & 45.3 & 87.8 & 88.5 & 7.4 & 2.5 & 33.8 \\
 & \cellcolor{LightGray}\textbf{ZoomCap-450K (Ours)} & \cellcolor{LightGray}\textbf{69.3} & \cellcolor{LightGray}\textbf{94.1} & \cellcolor{LightGray}\textbf{90.6} & \cellcolor{LightGray}\textbf{3.4} & \cellcolor{LightGray}\textbf{1.8} & \cellcolor{LightGray}\textbf{52.5} \\
\midrule
\multirow{3}{*}{Food}
 & ShareGPT4V-450K~\citep{chen2023sharegpt4v} & 22.4 & 81.7 & 80.7 & 15.9 & 2.2 & 15.9 \\
 & ScaleCap-450K~\citep{zhang2025scalecap} & 43.1 & 87.6 & 87.0 & 7.8 & 3.3 & 30.9 \\
 & \cellcolor{LightGray}\textbf{ZoomCap-450K (Ours)} & \cellcolor{LightGray}\textbf{74.0} & \cellcolor{LightGray}\textbf{94.1} & \cellcolor{LightGray}\textbf{92.1} & \cellcolor{LightGray}\textbf{3.0} & \cellcolor{LightGray}\textbf{1.5} & \cellcolor{LightGray}\textbf{58.6} \\
\midrule
\multirow{3}{*}{\makecell{People/\\animals}}
 & ShareGPT4V-450K~\citep{chen2023sharegpt4v} & 19.1 & 82.6 & 83.8 & 13.9 & \textbf{1.2} & 14.1 \\
 & ScaleCap-450K~\citep{zhang2025scalecap} & 41.8 & 90.7 & 89.0 & 7.1 & 1.9 & 32.4 \\
 & \cellcolor{LightGray}\textbf{ZoomCap-450K (Ours)} & \cellcolor{LightGray}\textbf{64.6} & \cellcolor{LightGray}\textbf{93.5} & \cellcolor{LightGray}\textbf{91.7} & \cellcolor{LightGray}\textbf{3.1} & \cellcolor{LightGray}1.6 & \cellcolor{LightGray}\textbf{52.1} \\
\bottomrule
\end{tabular}%
}
\end{table}


\begin{table}[htbp]
\centering
\scriptsize
\caption{\textbf{Two-anchor FactCycle quadrant shares by fact category.}
Shares are computed over claims decidable on both anchors within each
category: transferred (true and rendered), specious (fabricated yet
rendered), and unrendered fabrication (fabricated and unrendered); the
true-but-unrendered generator cell is excluded from caption metrics.
Shading \colorbox{myGreen!18}{\rule{0.9em}{0pt}\rule{0pt}{0.7ex}} marks the
best cell of each column (highest transferred, lowest specious and lowest
unrendered fabrication).}
\label{tab:dual_anchor_categories}
\resizebox{\textwidth}{!}{%
\begin{tabular}{llccccccc}
\toprule
\textbf{Caption source} & \textbf{Metric} & \textbf{entity} & \textbf{attribute} & \textbf{spatial} & \textbf{count} & \textbf{\makecell{small\\obj.}} & \textbf{action} & \textbf{other} \\
\midrule
\midrule
\multirow{3}{*}{ShareGPT4V-450K}
 & transferred (\%) $\uparrow$ & 89.5 & 78.8 & 77.4 & 68.0 & 78.7 & 70.7 & 86.7 \\
 & specious (\%) $\downarrow$ & 10.1 & 18.5 & 18.5 & 20.1 & 19.8 & 24.1 & 8.6 \\
 & unrendered fab. (\%) $\downarrow$ & \cellcolor{myGreen!18}0.1 & \cellcolor{myGreen!18}1.5 & 3.1 & 6.4 & \cellcolor{myGreen!18}0.9 & \cellcolor{myGreen!18}3.3 & 3.6 \\
\midrule
\multirow{3}{*}{ScaleCap-450K}
 & transferred (\%) $\uparrow$ & 94.2 & 87.8 & 82.8 & 83.8 & \cellcolor{myGreen!18}85.4 & 84.8 & 93.6 \\
 & specious (\%) $\downarrow$ & 4.8 & 8.2 & 9.4 & 7.0 & 8.4 & 8.1 & 4.0 \\
 & unrendered fab. (\%) $\downarrow$ & 0.5 & 2.4 & 4.5 & 4.9 & 3.7 & 4.2 & 1.8 \\
\midrule
\multirow{3}{*}{\textbf{ZoomCap-450K (Ours)}}
 & transferred (\%) $\uparrow$ & \cellcolor{myGreen!18}95.8 & \cellcolor{myGreen!18}91.0 & \cellcolor{myGreen!18}87.3 & \cellcolor{myGreen!18}84.8 & 84.2 & \cellcolor{myGreen!18}86.1 & \cellcolor{myGreen!18}94.6 \\
 & specious (\%) $\downarrow$ & \cellcolor{myGreen!18}1.6 & \cellcolor{myGreen!18}3.3 & \cellcolor{myGreen!18}4.9 & \cellcolor{myGreen!18}3.6 & \cellcolor{myGreen!18}5.9 & \cellcolor{myGreen!18}5.4 & \cellcolor{myGreen!18}2.4 \\
 & unrendered fab. (\%) $\downarrow$ & 0.6 & 1.7 & \cellcolor{myGreen!18}2.2 & \cellcolor{myGreen!18}3.2 & 3.0 & 3.8 & \cellcolor{myGreen!18}1.3 \\
\bottomrule
\end{tabular}%
}
\end{table}


\begin{table}[htbp]
\centering
\scriptsize
\caption{\textbf{Conservative dual-anchor accounting on FactCycle-250.}
In the conservative variant, reconstruction-leg unverifiable verdicts count
toward the unrealized denominator. Fabricated-then-rendered: share of claims
unsupported by the source image that the reconstruction nevertheless renders;
true-among-rendered: share of rendered claims that the source image also
supports; fabrication rate: share of claims unsupported by the source image
under the conservative denominator. Lower is better for all three.}
\label{tab:dual_anchor_conservative}
\begin{tabular}{lccc}
\toprule
\textbf{Caption source} & \textbf{\makecell{Fabricated-then-\\rendered (\%) $\downarrow$}} & \textbf{\makecell{True-among-\\rendered (\%) $\uparrow$}} & \textbf{\makecell{Fabrication\\rate (\%) $\downarrow$}} \\
\midrule
\midrule
ShareGPT4V-450K~\citep{chen2023sharegpt4v} & 88.1 & 79.7 & 18.4 \\
ScaleCap-450K~\citep{zhang2025scalecap} & 72.5 & 87.0 & 9.9 \\
\rowcolor{LightGray}
\textbf{ZoomCap-450K (Ours)} & \textbf{64.7} & \textbf{92.9} & \textbf{5.1} \\
\bottomrule
\end{tabular}
\end{table}

\paragraph{Category-level readings.}
ZoomCap keeps the lowest specious share in all seven fact categories and the
highest transferred share in six of them
(Table~\ref{tab:dual_anchor_categories}).
For ShareGPT4V-450K the fabrications concentrate on actions and counts, where
$24.1\%$ and $20.1\%$ of decidable claims are specious---fabricated
yet still rendered, so any reconstruction-only metric counts invented dynamics
and quantities as supported content.
ZoomCap's true-but-unrendered share (generator-attributed, excluded from caption metrics) is slightly elevated on counts and spatial
relations ($8.5\%$ and $5.6\%$), consistent with the fixed generator dropping
specific high-cardinality facts under dense captions; per the two-anchor
attribution this is generator loss rather than caption error, and it enters
$\mathrm{FC_{trans}}$ identically for all sources.
The unrendered-fabrication share likewise peaks in the count column for
ShareGPT4V ($6.4\%$) and ScaleCap ($4.9\%$) and, for ZoomCap, on actions
($3.8\%$, against $3.2\%$ on counts)---fabricated and unrenderable claims
concentrate where rendering is hardest, reinforcing the reading
that counting facts stress the fixed generator the most.

\paragraph{Realizability accounting.}
The reported $\rho$ drops reconstruction-unverifiable verdicts from both
sides (supported over decidable claims): 92.3/93.6/95.7\% for
ZoomCap/ScaleCap/ShareGPT4V. The stricter variant that charges unverifiable
verdicts to the caption (supported over all extracted claims) lowers these
rates to 88.8/90.7/93.4\%; the 3.5/2.9/2.3-point gaps are exactly each
corpus's unverifiable share, largest for ZoomCap because denser captions
offer more claims the fixed renderer cannot make checkable. Per-bucket
realizability (decidable-only) is highest for the shortest captions in every
scene bucket (ZoomCap 90.9--93.7, ScaleCap 92.2--94.7, ShareGPT4V
95.3--96.5), which is why realizability alone cannot rank caption quality;
the source ranking is identical under both accountings.

\paragraph{Judge configuration and robustness.}
Both anchors share the judge pair \texttt{Kimi-K3} and \texttt{Qwen3.8-Max};
per-claim exact agreement is 0.933 over the 32{,}205 claims both
realizability judges verdicted (64{,}617 judgments) and 0.877 over the
31{,}452 claims scored by both truthfulness judges.

\section{Corpus Statistics}
\label{sec:corpus_stats}

Table~\ref{tab:dataset_stats} gives the corpus side of the equal-scale
comparison run in Table~\ref{tab:cpt_comparison}: caption length, entity
density, and the image source of each 450K set.


\begin{center}
\scriptsize
\captionsetup{type=table}
\caption{\textbf{Caption-corpus statistics for equal-scale 450K datasets.} ZoomCap-450K contains more text and named/object entities; downstream utility is evaluated in Table~\ref{tab:cpt_comparison}.}
\label{tab:dataset_stats}
\resizebox{\textwidth}{!}{%
\begin{tabular}{lccccccc}
\toprule
\textbf{Dataset} & \textbf{\#Images} & \textbf{Image source} & \textbf{Caption engine} & \textbf{Avg. chars} & \textbf{Avg. words} & \textbf{Avg. sents} & \textbf{Avg. entities} \\
\midrule
ShareGPT4V-450K~\citep{chen2023sharegpt4v} & 450K & ShareGPT4V-100K + sampled ShareGPT4V-PT & GPT-4V / Share-Captioner & 850.7 & 147.6 & 8.4 & 2.3 \\
DenseFusion-450K~\citep{li2024densefusion} & 450K & Sampled from DenseFusion-1M & DenseFusion & 1129.1 & 193.5 & 10.5 & 3.8 \\
ScaleCap-450K~\citep{zhang2025scalecap} & 450K & ShareGPT4V-100K + LAION-5B subset~\citep{schuhmann2021laion} & ScaleCap & 2907.8 & 462.2 & 21.5 & 5.8 \\
\rowcolor{LightGray}
\textbf{ZoomCap-450K (Ours)} & 450K & ShareGPT4V-100K + LAION-5B subset~\citep{schuhmann2021laion} & ZoomCap (Qwen3.5-35B-A3B) & \textbf{4228.6} & \textbf{622.8} & \textbf{33.1} & \textbf{12.4} \\
\bottomrule
\end{tabular}%
}
\end{center}

\section{General-Domain Adoption Details}
\label{sec:adoption_curves}

\begin{figure}[!htbp]
\centering
\includegraphics[width=0.8\textwidth]{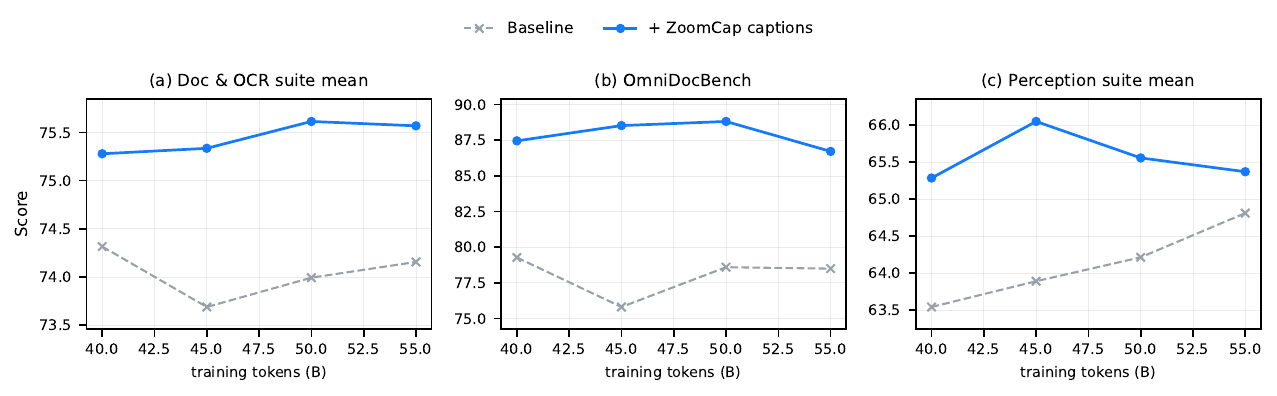}
\caption{\textbf{Checkpoint-level curves for the general-domain adoption study.} Suite means over the shared checkpoints (40--55B tokens) for the baseline mixture and the mixture with ZoomCap captions: (a)~the nine-benchmark Doc \& OCR suite, (b)~OmniDocBench, (c)~the nine-benchmark perception suite; the gains are stable across checkpoints.}
\label{fig:adoption_curves}
\end{figure}

Table~\ref{tab:adoption_general} compares two matched production runs of
\texttt{Ling-Tiny-VL-v3} under a common 60B-token budget: they share checkpoint,
hyperparameters, and seed; the experimental arm swaps the production
\texttt{vl\_caption} snapshot (\texttt{pretrain\_caption\_20260623}) for
\texttt{zoomcap\_pretrain\_1600w}---16M ZoomCap captions up-sampled $4\times$ to
60M and mixed with 240M base captions so that ZoomCap forms $20\%$ of the
caption group by count---and every sub-score averages the four checkpoints
(40--55B tokens) common to both runs. The gain concentrates in the
document/OCR suite ($+1.4$ on the nine-benchmark mean, driven by OmniDocBench
$+9.8$ and OCRBenchV2-b $+2.2$) and is comparably large on the perception
suite ($+1.5$, led by MMVet $+11.8$ and HallusionBench $+1.4$) and the
knowledge-VQA suite ($+0.7$). Figure~\ref{fig:adoption_curves} reports the same
suite means per checkpoint instead of averaged over them, so the
document-structure and perception gains read as stable across the shared
checkpoints rather than as a single-checkpoint artifact.

\section{Medical-Domain Adoption Details}
\label{sec:medical_adoption}

In the medical-domain study (Table~\ref{tab:adoption_medical} in the main paper), the reference recipe is itself a medical-oriented pretraining mixture: the dedicated medical-image set (\texttt{20260731\_medical\_image}, 3.66M images) and interleaved medical documents together account for 50\% of the mixture.
We introduce 2.64M medical-image captions as a dedicated group covering 20\% of the mixture, rebalanced from the interleaved medical-document group; the reference recipe spends the same medical budget without this caption group, and the same image set is captioned either by Gemini-3.1-Pro~\citep{google2026gemini31pro} single-pass or by ZoomCap, so the two augmented runs isolate the caption engine under an otherwise identical recipe.
Concretely, both augmented runs keep the original 3.66M-image stream and add the dedicated group on top: the 2.64M engine captions enter the mixture as 15.55M caption samples, expanding the medical-image stream from 3.66M samples (1.73B tokens) to 19.21M samples (8.58B tokens).
ZoomCap captions improve the five medical benchmarks by +3.04 on average, versus +1.76 for single-pass Gemini-3.1-Pro captions, and turn VQA-RAD~\citep{lau2018vqarad} from a slight regression into +3.1.
ZoomCap also has the smaller change on the internal hallucination monitor ($-0.5$
vs.\ $-1.5$ on hallusion\_intern); this diagnostic does not by itself identify the
cause of the difference.
Both augmented runs train on 64 GPUs for $\sim$50B tokens against the 128-GPU, $\sim$100B-token baseline: at half the compute budget, both caption recipes outperform the reference, and ZoomCap outperforms the stronger captioning engine.

\section{Efficiency and Auditability}
\label{sec:efficiency}

In practice, the dominant cost is the number of observations rather than any single local call.
This explains why budgeted zooming is part of the method: reducing unnecessary second-level patches improves efficiency while also reducing the chance of local hallucination.
For token-metered systems, the normalized per-image cost is
\begin{equation}
    \mathrm{Cost}_{\mathrm{image}}
    =
    \sum_{s\in\mathcal{S}} n_s
    \left(
    \frac{\mathbb{E}[t^{\mathrm{in}}_s]}{10^6}P_{\mathrm{in}}
    +
    \frac{\mathbb{E}[t^{\mathrm{out}}_s]}{10^6}P_{\mathrm{out}}
    \right),
\end{equation}
where $n_s$ is the number of calls at stage $s$, $t^{\mathrm{in}}_s$ and $t^{\mathrm{out}}_s$ are token counts, and $P_{\mathrm{in}},P_{\mathrm{out}}$ are prices per million tokens.

Table~\ref{tab:cost_analysis} instantiates this accounting for the engines.
Measured over our production logs, the full ZoomCap evidence workflow costs \$0.0077 per image---of the same order as the logged single-pass Qwen3.5-397B-A17B~\citep{qwen2026qwen35} call, and $3.6\times$ cheaper than the official ScaleCap~\citep{zhang2025scalecap} recipe measured on the same service and teacher, while transferring $1.66\times$ more claims (Table~\ref{tab:dual_anchor}); equivalently, each transferred claim costs about $\$1.5\times10^{-4}$ for ZoomCap versus $\$8.7\times10^{-4}$ for ScaleCap, i.e., ${\sim}5.9\times$ cheaper per unit of transferred evidence. The gap is robust to the remaining modeling assumptions: even halving ScaleCap's measured input tokens would leave it ${\sim}2.0\times$ more expensive per image than ZoomCap. At this rate, captioning the 2.64M medical images of Section~\ref{sec:adoption} costs about \$20K.


\begin{center}
\scriptsize
\setlength{\tabcolsep}{4pt}
\captionsetup{type=table}
\caption{\textbf{Per-image captioning cost.} All rows caption one image at $768$ pixels with a single teacher family; ZoomCap is measured from production workflow logs of the Qwen3.5-35B-A3B teacher, the ScaleCap row by running its official recipe and code~\citep{zhang2025scalecap} through the identical asynchronous service at the same teacher, and the Qwen3.5-397B-A17B row from the logged single-pass flagship run that supplies the PrefBench-456 opposing captions. Token prices are \$0.25/\$0.60 per $10^6$ (Qwen3.5-35B-A3B) and \$1.20/\$6.00 (Qwen3.5-397B-A17B); output tokens count every generated text (drafts, evidence/question--answer streams, patch captions, final caption). Wall-clock is per image in the asynchronous service (ZoomCap: concurrent knowledge/grounding stages, five parallel patch workers; ScaleCap: heuristic rounds on the same workers). The final column divides \$/img by $\mathrm{FC_{trans}}$ transferred claims (Table~\ref{tab:dual_anchor}); single-pass rows are blank as those captions were not scored under that protocol.}
\label{tab:cost_analysis}
\renewcommand{\arraystretch}{0.85}
\resizebox{\textwidth}{!}{%
\begin{tabular}{@{}lrrrrrrr@{}}
\toprule
\textbf{Captioning engine} & \textbf{Teacher} & \textbf{Calls/img} & \textbf{Input tok/img} & \textbf{Output tok/img} & \textbf{\$/img} & \textbf{Wall-clock (s)} & \textbf{\$/transferred claim} \\
\midrule
\midrule
Single-pass draft & Qwen3.5-35B-A3B~\citep{qwen2026qwen35} & 1 & 1.3K & 0.4K & 0.0006 & 4.6 & -- \\
Single-pass flagship & Qwen3.5-397B-A17B~\citep{qwen2026qwen35} & 1 & 2.1K & 0.7K & 0.0067 & $\sim$17 & -- \\
ScaleCap-450K recipe~\citep{zhang2025scalecap} & Qwen3.5-35B-A3B & $\sim$70 & 93.6K & 6.6K & 0.0274 & $\sim$123 & $8.7\times10^{-4}$ \\
\midrule
\rowcolor{LightGray}
\textbf{ZoomCap (Ours)} & Qwen3.5-35B-A3B & $\sim$9 & 20.6K & 4.3K & \textbf{0.0077} & $\sim$29 & $\mathbf{1.5\times10^{-4}}$ \\
\bottomrule
\end{tabular}%
}
\end{center}

\paragraph{Concurrency in the serving pipeline.}
The workflow is implemented as an asynchronous FastAPI service: each image is an independent job served by a configurable worker pool.
Once the global draft completes, the knowledge and grounding stages are dispatched concurrently, the local patch calls run in parallel over five workers, and distinct images are processed simultaneously, so per-image wall-clock follows the critical path rather than the sum of calls (the $\sim$29 s of Table~\ref{tab:cost_analysis}, against the roughly 55 s that the same calls would take sequentially).

Because every stage writes intermediate evidence artifacts, the workflow is also auditable: failures can be traced to routing, search context, OCR, local patches, relation facts, or fusion rather than being hidden inside one opaque generation call.

\end{document}